\documentclass{article}

\PassOptionsToPackage{numbers, sort&compress}{natbib}

\usepackage{microtype}
\usepackage{algorithm}
\usepackage{graphicx}
\usepackage{subcaption}
\usepackage{booktabs} %
\usepackage{xcolor}
\usepackage{bbm}
\usepackage{todonotes}
\usepackage{cancel}
\usepackage{amsthm}
\usepackage{wrapfig}
\usepackage[noend]{algpseudocode}
\usepackage{enumitem}%
\algrenewcommand\algorithmicindent{1em}

\definecolor{custom_blue}{rgb}{0.12156863, 0.46666667, 0.70588235}
\definecolor{custom_orange}{rgb}{1., 0.5, 0.05}

\usepackage{amsmath,amsfonts,bm}

\newcommand{\figleft}{{\em (Left)}}
\newcommand{\figcenter}{{\em (Center)}}
\newcommand{\figright}{{\em (Right)}}

\def\eqref#1{equation~\ref{#1}}

\def\1{\bm{1}}

\DeclareMathAlphabet{\mathsfit}{\encodingdefault}{\sfdefault}{m}{sl}
\SetMathAlphabet{\mathsfit}{bold}{\encodingdefault}{\sfdefault}{bx}{n}

\def\gA{{\mathcal{A}}}

\def\gM{{\mathcal{M}}}

\def\gQ{{\mathcal{Q}}}

\def\gS{{\mathcal{S}}}

\newcommand{\E}{\mathbb{E}}

\DeclareMathOperator*{\argmax}{arg\,max}

\usepackage{amssymb}
 
\newtheorem{lemma}{Lemma}

\usepackage{mathtools}

\DeclarePairedDelimiterX{\infdivx}[2]{(}{)}{%
  #1\;\delimsize\|\;#2%
}
\newcommand{\kl}{D_{\mathrm{KL}}\infdivx}

\usepackage[final]{neurips_2019}
\usepackage{hyperref}

\title{Rewriting History with Inverse RL: \\Hindsight Inference for Policy Improvement}

\author{%
  Benjamin Eysenbach\thanks{Equal contribution. Correspondence to \href{mailto:beysenba@cs.cmu.edu}{beysenba@cs.cmu.edu}}$^{\phantom{*}\phi\theta}$ %
  \quad Xinyang Geng$^{*\psi}$
  \quad Sergey Levine$^{\psi\theta}$
  \quad Ruslan Salakhutdinov$^{\phi}$ \\
  $^\phi$ Carnegie Mellon University
  \qquad $^\psi$ UC Berkeley
  \qquad $^\theta$ Google Brain
}
\begin{document}
\begin{NoHyper}
\maketitle
\end{NoHyper}

\begin{abstract}
Multi-task reinforcement learning (RL) aims to simultaneously learn policies for solving many tasks.
Several prior works have found that relabeling past experience with different reward functions can improve sample efficiency. Relabeling methods typically ask: if, in hindsight, we assume that our experience was optimal for some task, for what task was it optimal? In this paper, we show that \emph{hindsight relabeling is inverse RL}, an observation that suggests that we can use inverse RL in tandem for RL algorithms to efficiently solve many tasks. We use this idea to generalize goal-relabeling techniques from prior work to arbitrary classes of tasks. Our experiments confirm that relabeling data using inverse RL accelerates learning in general multi-task settings, including goal-reaching, domains with discrete sets of rewards, and those with linear reward functions.

\end{abstract}

\vspace{-0.5em}
\section{Introduction}
\vspace{-0.5em}
\label{sec:intro}

Reinforcement learning (RL) aims to acquire control policies that take actions to maximize their cumulative reward. Existing RL algorithms remain data inefficient, requiring exorbitant amounts of experience to learn even simple tasks (e.g., ~\citep{dubey2018investigating, kapturowski2018recurrent}).
Multi-task RL, where many RL problems are solved in parallel, has the potential to be more sample efficient than single-task RL, as data can be shared across tasks.
Nonetheless, the problem of effectively sharing data across tasks remains largely~unsolved.

The idea of sharing data across tasks has been studied at least since the 1990s~\citep{caruana1997multitask}.
More recently, a number of works have observed that retroactive relabeling of experience with different tasks can improve data efficiency.
A common theme in prior relabeling methods is to relabel past trials with whatever goal or task was performed successfully in that trial. For example, relabeling for a goal-reaching task might use the state actually reached at the end of the trajectory as the relabeled goal, sine the trajectory corresponds to a successful trial \emph{for the goal that was actually reached}~\citep{kaelbling1993learning, andrychowicz2017hindsight, pong2018temporal}. However, prior work has presented these goal-relabeling methods primarily as heuristics, and it remains unclear how to intelligently apply the same idea to tasks other than goal-reaching, such as those with linear reward functions.

\begin{figure}[t]
    \centering
    \vspace{-1.5em}
    \includegraphics[width=0.8\linewidth]{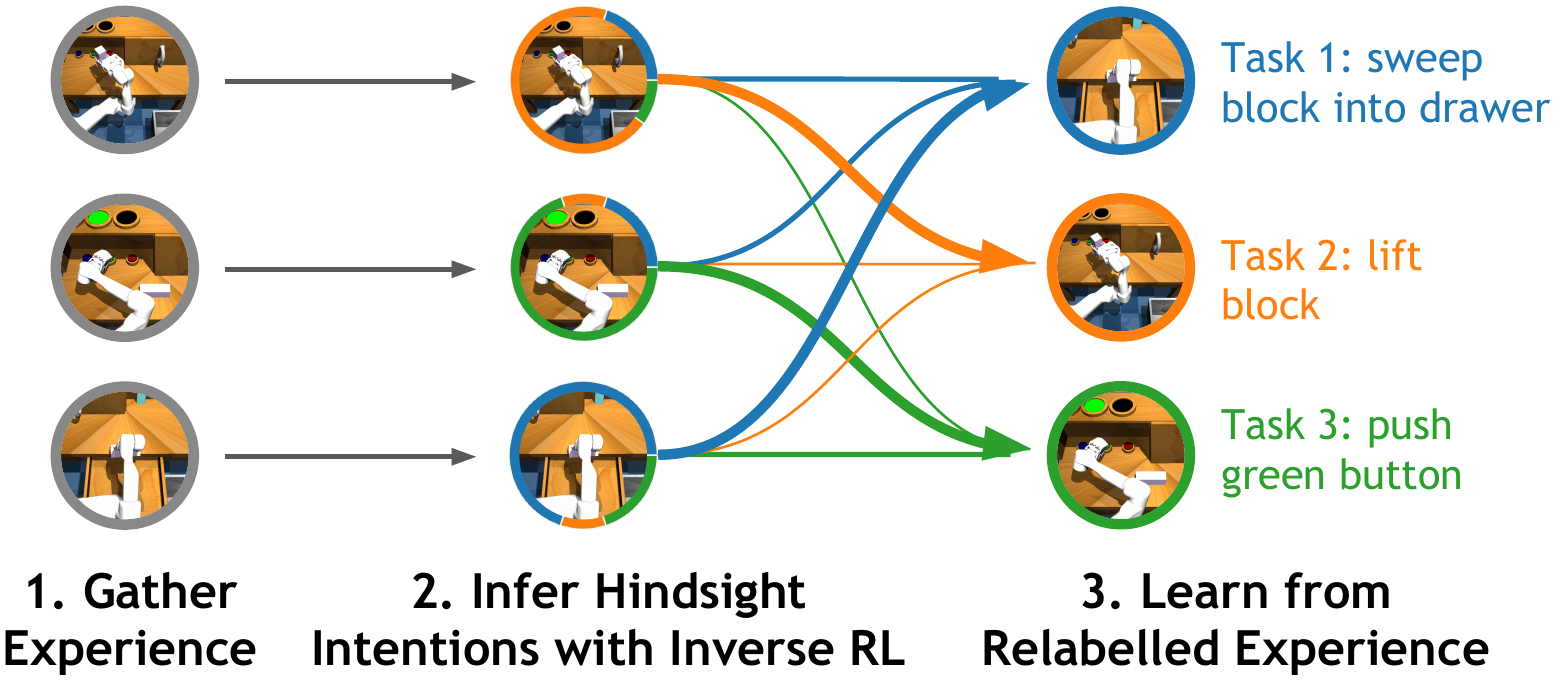}
    \caption{\textbf{Hindsight Inference for Policy Improvement (HIPI)}:
    Given a dataset of prior experience, we use inverse RL to infer the agent's intentions. We use the relabeled experience with any policy learning algorithm, such as off-policy RL or supervised learning. \label{fig:teaser}}
    \vspace{-1.5em}
\end{figure}
In this paper, we formalize prior relabeling techniques under the umbrella of \emph{inverse} RL: by inferring the most likely task for a given trial via inverse RL, we provide a principled formula for relabeling in arbitrary multi-task problems.
Inverse RL is \emph{not} the same as simply assigning each trajectory to the task for which it received the highest reward. In fact, this strategy would often result in assigning most trajectories to the easiest task. Rather, inverse RL takes into account the difficulty of different tasks and the amount of reward that each yields. RL and inverse RL can be seen as complementary tools for maximizing reward: RL takes tasks and produces high-reward trajectories, and inverse RL takes trajectories and produces task labels such that the trajectories receive high reward. Formally, we prove that maximum entropy (MaxEnt) RL and MaxEnt inverse RL optimize the same multi-task objective: MaxEnt RL optimizes with respect to trajectories, while MaxEnt inverse RL optimizes with respect to tasks.
Unlike prior goal-relabeling techniques, we can use inverse RL to relabel experience for arbitrary task distributions, including sets of linear or discrete rewards.  
This observation suggests that tools from RL and inverse RL might be combined to efficiently solve many tasks simultaneously.
The combination we develop, Hindsight Inference for Policy Improvement (HIPI), first relabels experience with inverse RL and then uses the relabeled experience to learn a policy (see Fig.~\ref{fig:teaser}). 
One variant of this framework follows the same design as prior goal-relabeling methods~\citep{kaelbling1993learning, andrychowicz2017hindsight, pong2018temporal} but uses inverse RL to relabel experience, a difference that allows our method to handle arbitrary task families. The second variant has a similar flavour to self-imitation behavior cloning methods~\citep{oh2018self,ghosh2019learning, savinov2018semi}: we relabel past experience using inverse RL and then learn a policy via task-conditioned behavior cloning.
Both algorithms can be interpreted as probabilistic reinterpretation and generalization of prior work.

The main contribution of our paper is the observation that hindsight relabeling is inverse RL. This observation not only provides insight into success of prior relabeling methods, but it also provides guidance on applying relabeling to arbitrary multi-task RL problems. That RL and inverse RL can be used in tandem is not a coincidence; we prove that MaxEnt RL and MaxEnt inverse RL optimize the same multi-task RL objective with respect to trajectories and tasks, respectively. Our second contribution consists of two simple algorithms that use inverse RL-based relabeling to accelerate RL. %
Our experiments on complex simulated locomotion and manipulation tasks demonstrate that our method outperforms state-of-the-art methods on tasks ranging from goal-reaching, running in various directions, and performing a host of manipulation tasks.

\vspace{-0.5em}
\section{Prior Work}
\vspace{-0.5em}
\label{sec:related-work}

The focus of our work is on multi-task RL problems, for which a number of algorithms have been proposed over the past decades~\citep{thrun2012learning, hessel2019multi, teh2017distral, espeholt2018impala, riedmiller2018learning}.
Existing approaches still struggle to reuse data across multiple tasks, with researchers often finding that training separate models is a very strong baseline~\citep{yu2020gradient} and using independently-trained models as an initialization or prior for multi-task models~\citep{parisotto2015actor, rusu2015policy, ghosh2017divide, teh2017distral}.
When applying off-policy RL in the multi-task setting, a common trick is to take experience collected when performing task A and pretend that it was collected for task B by recomputing the rewards at each step. This technique effectively inflates the amount of data available for learning, and a number of prior works have found this technique quite effective~\citep{kaelbling1993learning, pong2018temporal, andrychowicz2017hindsight, schaul2015universal}.
In this paper we show that the relabeling done in prior work can be understood as inverse RL.

If RL is asking the question of how to go from a reward function to a policy, inverse RL asks the opposite question: after observing an agent acting in an environment, can we infer which reward function the agent was trying to optimize? A number of inverse RL algorithms have been proposed~\citep{ratliff2006maximum, abbeel2004apprenticeship}, with MaxEnt inverse RL being one of the most commonly used frameworks~\citep{ziebart2008maximum, finn2016guided, javdani2015shared}. Since MaxEnt inverse RL can be viewed as an inference problem, we can calculate either the posterior distribution over reward functions, or the maximum a-posteriori (MAP) estimate. While most prior work is concerned with MAP estimates, we follow~\citet{hadfield2017inverse} in using the full posterior distribution. Section~\ref{sec:prelim-irl} discusses how MaxEnt RL and MaxEnt inverse RL are closely connected,
with one problem being the dual of the other. It is therefore not a coincidence that many MaxEnt inverse RL algorithms involve solving a MaxEnt RL problem in the inner loop. Our paper proposes the opposite, using MaxEnt inverse RL in the inner loop of MaxEnt RL.

Our work builds on the idea that MaxEnt RL can be viewed as probabilistic inference. This idea has been proposed in a number of prior works~\citep{kappen2012optimal, toussaint2009robot, todorov2008general, todorov2007linearly, rawlik2013stochastic, theodorou2012relative, levine2018reinforcement} and used to build a number of modern RL algorithms~\citep{haarnoja2017reinforcement, haarnoja2018soft, abdolmaleki2018maximum}. Perhaps the most relevant prior work is~\citet{rawlik2013stochastic}, which emphasizes that MaxEnt RL can be viewed as minimizing an KL divergence, an idea that we extend to the multi-task setting.

\vspace{-0.5em}
\section{Preliminaries}
\vspace{-0.5em}
\label{sec:prelim}

This section reviews MaxEnt RL and MaxEnt inverse RL.
We start by introducing notation.

\vspace{-0.5em}
\paragraph{Notation}
We will analyze an MDP $\gM$ with states $s_t \in \gS$ and reward function $r(s_t, a_t)$. We assume that actions $a_t \in \gA$ are sampled from a policy $q(a_t \mid s_t)$. The initial state is sampled $s_1 \sim p_1(s_1)$ and subsequent transitions are governed by a dynamics distribution $s_{t+1} \sim p(s_{t+1} \mid s_t, a_t)$. We define a trajectory as a sequence of states and actions: $\tau = (s_1, a_1, \cdots)$, and write the likelihood of a trajectory under policy $q$ as
\begin{equation}
    q(\tau) = p_1(s_1) \prod_t p(s_{t+1} \mid s_t, a_t) q(a_t \mid s_t).
\end{equation}
In the multi-task setting, we will use $\psi \in \Psi$ to identify each task, and assume that we are given a prior $p(\psi)$ over tasks. The set of tasks $\Psi$ can be continuous or discrete, finite or infinite; each particular task $\psi \in \Psi$ can be continuous or discrete valued. We define $r_\psi(s_t, a_t)$ as the reward function for task $\psi$. Our experiments will use both goal-reaching tasks, where $\psi$ is a goal state, as well as more general task distributions, where $\psi$ is the hyperparameters of the reward function

\vspace{-0.5em}
\paragraph{MaxEnt RL}
MaxEnt RL casts the RL problem as one of sampling trajectories with probability proportional to exponentiated reward. Given a reward function $r(s_t, a_t)$, we aim to learn a policy that samples trajectories from the following target distribution, $p(\tau)$:
\begin{align}
    p(\tau) &\triangleq \frac{1}{Z} p_1(s_1) \prod_t p(s_{t+1} \mid s_t, a_t) e^{r(s_t, a_t)}. \label{eq:maxent-likelihood}
\end{align}
The partition function $Z$ is introduced to make $p(\tau)$ integrate to one.
The objective function for MaxEnt RL is to maximize the entropy-regularized sum of rewards, which is equivalent to minimizing the reverse KL divergence between the policy's distribution over trajectories, $q(\tau)$, and a target distribution, $p(\tau)$ defined in terms of rewards $r_t = r(s_t, a_t)$:
\begin{align*}
-\kl{q}{p} = \E_q\left[\left(\sum_t r_t - \log q(a_t \mid s_t)\right) - \log Z\right].
\end{align*}
The partition function does not depend on the policy, so prior RL algorithms have ignored it.

\vspace{-0.5em}
\paragraph{MaxEnt Inverse RL}
\label{sec:prelim-irl}
Inverse RL observes previously-collected data and attempts to infer the intent of the actor, which is represented by a reward function $r_\psi$. MaxEnt inverse RL is a variant of inverse RL that defines the probability of trajectory $\tau$ being produced for task $\psi$ as
\begin{equation*}
p(\tau \mid \psi) = \frac{1}{Z(\psi)} p_1(s_1) \prod_t p(s_{t+1} \mid s_t, a_t) e^{r_\psi(s_t, a_t)},
\end{equation*}
where
\begin{equation*}
     Z(\psi) \triangleq \int p_1(s_1) \prod_t p(s_{t+1} \mid s_t, a_t) e^{r_\psi(s_t, a_t)} d \tau. \label{eq:multi-task-partition}
\end{equation*}
Applying Bayes' Rule, the posterior distribution over reward functions is given as follows:
\begin{equation}
    p(\psi \mid \tau) = \frac{p(\tau \mid \psi) p(\psi)}{p(\tau)} \propto p(\psi) e^{\sum_t r_\psi(s_t, a_t) - \log Z(\psi)}. \label{eq:irl}
\end{equation}
While many applications of MaxEnt inverse RL use the maximum a posteriori estimate, $\argmax_\psi \left[ \log p(\psi \mid \tau) \right]$ in this paper will use the full posterior distribution. %
While the partition function, an integral over all states and actions, is typically hard to compute, its dual is the MaxEnt RL problem:
\begin{equation}
     \log Z(\psi) = \max_{q(\tau \mid \psi)} \E_{q(\tau \mid \psi)} \! \bigg[\sum_t r_\psi(s_t, a_t) - \log q(a_t \mid s_t, \psi)\bigg].
\end{equation}
The striking similarities between MaxEnt RL and MaxEnt inverse RL are not a coincidence. As we will show in the next section, both minimize the same reverse KL divergence on the joint distribution of tasks and trajectories.

\section{Hindsight Relabeling is Inverse RL}
\label{sec:multi-task}

We now aim to use the tools of RL and inverse RL to solve many RL problems simultaneously, each with the same dynamics but a different reward function. 
Given a prior over tasks, $p(\psi)$, the target joint distribution over tasks and trajectories is
\begin{align}
    p(\tau, \psi) = p(\psi) \frac{1}{Z(\psi)} p_1(s_1) \prod_t p(s_{t+1} \mid s_t, a_t) e^{r_\psi(s_t, a_t)}. \label{eq:target-dist}
\end{align}
We can express the multi-task (MaxEnt) RL objective as the reverse KL divergence between the joint trajectory-task distributions:
\begin{equation}
    \max_{q(\tau, \psi)} -\kl{q(\tau, \psi)}{p(\tau, \psi)}. \label{eq:kl-joint}
\end{equation}
If we factor the joint distribution as $q(\tau, \psi) = q(\tau \mid \psi) p(\psi)$, Eq.~\ref{eq:kl-joint} is equivalent to maximizing the expected (entropy-regularized) reward of a task-conditioned policy $q(\tau \mid \psi)$:
\begin{equation*}
    \E_{\substack{\psi \sim q(\psi) \\ \tau \sim  q(\tau \mid \psi)}} \left[ \left(\sum_r r_\psi(s_t, a_t) - \log q(a_t \mid s_t, \psi)\right) -\cancel{\log Z(\psi)} \right].
\end{equation*}
Since the distribution over tasks, $p(\psi)$ is fixed, we can ignore the $\log Z(\psi)$ term for optimization. 
A less common but more intriguing choice is to factor $q(\tau, \psi) = q(\psi \mid \tau) q(\tau)$, where $q(\tau)$ is represented non-parametrically as a distribution over previously-observed trajectories, and $q(\psi \mid \tau)$ is a \emph{relabeling distribution}. We find the optimal relabeling distribution by first rewriting~Eq.~\ref{eq:kl-joint}

\begin{align*}
    \E_{\substack{\tau \sim q(\tau) \\ \psi \sim  q(\psi \mid \tau)}} \bigg[&\cancel{\log p_1(s_1)} + \sum_t r_\psi(s_t, a_t) + \cancel{\log p(s_{t+1} \mid s_t, a_t)} \\
    & + \underbrace{p(\psi) - \log q(\psi \mid \tau)}_{-\kl{q(\psi \mid \tau)}{p(\psi)}} - \cancel{\log q(\tau)} - \log Z(\psi) \bigg],
\end{align*}
and then solving for the optimal relabeling distribution, ignoring terms that do not depend on $\psi$:
\begin{equation}
    q(\psi \mid \tau) \propto p(\psi) e^{\sum_t r_\psi(s_t, a_t) - \log Z(\psi)}. \label{eq:opt-relabel}
\end{equation}

The key observation here is that \emph{the optimal relabeling distribution corresponds exactly to MaxEnt inverse RL posterior over tasks} (Eq.~\ref{eq:irl}). Thus, we can obtain the optimal relabeling distribution via inverse RL.
While the optimal relabeling distribution derived here depends on the entire trajectory, Appendix~\ref{appendix:soft-q} shows how to perform relabeling when given a transition rather than an entire trajectory:
\begin{equation}
    q(\psi \mid s_t, a_t) \propto p(\psi) e^{\widetilde{Q}^q(s_t, a_t) - \log Z(\psi)} \label{eq:opt-relabel-transition}
\end{equation}
In the next section we show that prior goal-relabeling methods are a special case of inverse RL.

\subsection{Special Case: Goal Relabeling}
\label{sec:special-case}
A number of prior works have explicitly~\citep{kaelbling1993learning, andrychowicz2017hindsight, pong2018temporal} and implicitly~\citep{savinov2018semi,lynch2019learning,ghosh2019learning} found that hindsight relabeling can accelerate learning for \emph{goal-reaching} tasks, where tasks $\psi$ correspond to goal states. These prior relabeling methods are a special case of inverse RL.  We define a goal-conditioned reward function that penalizes the agent for failing to reaching the goal at the terminal step:
\begin{equation}
    r_\psi(s_t, a_t) = \begin{cases}
    -\infty & \text{ if } t = T \text{ and } s_t \neq \psi \\
    0 & \text{ otherwise}
    \end{cases}. \label{eq:neg-inf-rew}
\end{equation}
We assume that the time step $t$ is included in the observation $s_t$ to ensure that this reward function is Markovian. With this reward function, the optimal relabeling distribution $q(\psi \mid \tau)$ from Eq.~\ref{eq:opt-relabel} is~simply $q(\psi \mid \tau) = \mathbbm{1}(\psi = s_T)$,
where $s_T$ is the final state in trajectory $\tau$. Thus, \emph{relabeling with the state actually reached is equivalent inverse RL when using the reward function in Eq.~\ref{eq:neg-inf-rew}.}
While inverse RL is particularly convenient when using this reward function, it is rarely the metric of success that we actually care about. Viewing goal relabeling as a special case of inverse RL under a special reward function allows us to extend goal relabeling to general task arbitrary reward functions and arbitrary task distributions.
In our experiments, we show that inverse RL seamlessly handles task distributions including goal-reaching, discrete sets of tasks, and linear reward functions. 

\subsection{The Importance of the Partition Function}

\begin{wrapfigure}[14]{r}{0.5\textwidth}
\centering
\vspace{-2.5em}
\includegraphics[width=0.9\linewidth]{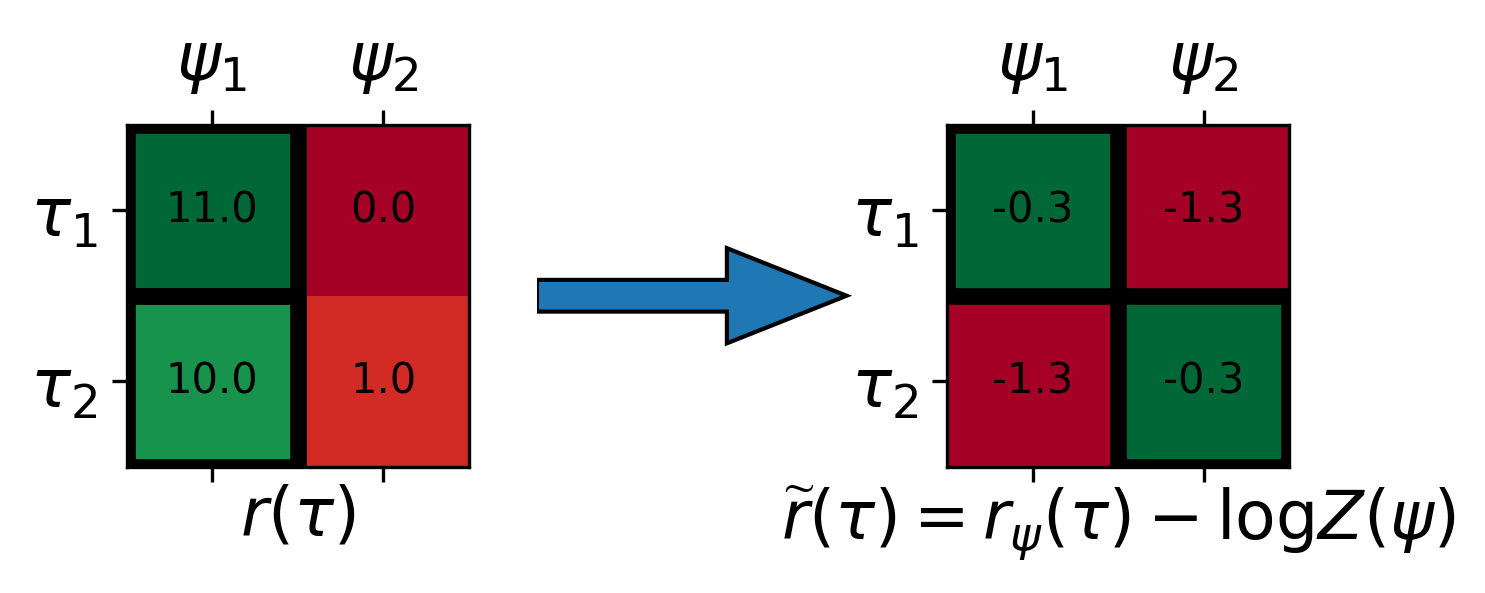}
\vspace{-0.5em}
\caption{\textbf{The partition function normalizes rewards of different scales}: {\footnotesize Two trajectories are evaluated on tasks with different reward scales. Black borders indicate the task to which we assign each trajectory. \figleft \; Without normalization, both trajectories are assigned to task $\psi_1$. \figright \; After normalizing with the partition function, as is done by inverse RL (our method), trajectory $\tau_1$ is assigned task $\psi_1$ and $\tau_2$ is assigned to~$\psi_2$. \label{fig:partition-fn-toy}}}
\end{wrapfigure}

The partition function used by inverse RL will be important for hindsight relabeling, as it will normalize the rewards from tasks with varying difficulty and reward scale. Fig.~\ref{fig:partition-fn-toy} shows a didactic example with two tasks, where the rewards for one task are larger than the rewards for the other task. Relabeling with the reward under which the agent received the largest reward (akin to~\citet{andrychowicz2017hindsight}) fails, because all experience will be relabeled with the first (easier) task.
Subtracting the partition function from the rewards (as in Eq.~\ref{eq:opt-relabel})
results in the desired behavior, trajectory $\tau_1$ is assigned task $\psi_1$ and $\tau_2$ is assigned to $\psi_2$.

\setlength{\textfloatsep}{0pt}%
\setlength{\floatsep}{0pt}%

\subsection{How Much Does Relabeling Help?}
Up to now, we have shown that the optimal way to relabel data is via inverse RL. How much does relabeling help? We now obtain a lower bound on the improvement from relabeling. Both lemmas in this section will assume that a joint distribution $q(\tau, \psi)$ over tasks and trajectories be given (e.g., specified by a policy $q(a_t \mid s_t, \psi)$). We will define $q_\tau(\tau) = \int q(\tau, \psi)$ as the marginal distribution over trajectories and then construct $q_\tau(\tau, \psi) = q_\tau(\psi \mid \tau) q_\tau(\tau)$ using the optimal relabeling distribution $q_\tau(\psi \mid \tau)$ (Eq.~\ref{eq:opt-relabel}).
We first show that relabeling data using inverse RL improves the MaxEnt RL objective: %
\begin{lemma}
The relabeled distribution $q_\tau(\tau, \psi)$ is closer to the target distribution than the original distribution, as measured by the KL divergence:
\begin{equation*}
    \kl{q_\tau(\tau, \psi)}{p(\tau, \psi)} \le \kl{q(\tau, \psi)}{p(\tau, \psi)}.
\end{equation*}
\end{lemma}
\vspace{-1.0em}
\begin{proof}
Of the many possible relabeling distributions, one choice is to do no relabeling, assigning to each trajectory $\tau$ the task $\psi$ that was commanded when the trajectory was collected. Denote this relabeling distribution $q_0(\psi \mid \tau)$, so $q_0(\psi \mid \tau) q_\tau(\tau) = q(\tau, \psi)$.
Because $q_\tau(\psi \mid \tau)$ was chosen as that which minimizes the KL among all relabeling distributions (including $q_0(\psi \mid \tau)$), the desired inequality holds:
\begin{align*}
    \kl{q_\tau(\psi \mid \tau) q_\tau(\tau)}{p(\tau, \psi)}  \le \kl{q_0(\psi \mid \tau) q_\tau(\tau)}{p(\tau, \psi)}.%
\end{align*}
\vspace*{-\baselineskip }\qedhere
\end{proof}
Thus, the relabeled data is an improvement over the original data, achieving a larger entropy-regularized reward (Eq.~\ref{eq:kl-joint}). As our experiments will confirm, relabeling data will accelerate learning.
Our next result will give us a lower bound on this improvement:
\begin{lemma} \label{lemma:improvement}
The improvement in the MaxEnt RL objective (Eq.~\ref{eq:kl-joint}) gained by relabeling is lower bounded as follows:
\begin{align*}
    & \kl{q(\tau, \psi)}{p(\tau, \psi)} - \kl{q_\tau(\tau, \psi)}{p(\tau, \psi)} \ge \E_{q_\tau} \left[\kl{q(\psi \mid \tau)}{q_\tau(\psi \mid \tau)} \right].
\end{align*}
\end{lemma}
The proof, a straightforward application of information geometry, is in Appendix~\ref{appendix:proof}. This result says that the amount that relabeling helps is at least as large as the difference between the task labels \mbox{$q(\psi \mid \tau)$} and the task labels inferred by inverse RL, $q_\tau(\psi \mid \tau)$. Note that, when we have learned the optimal policy (Eq.~\ref{eq:target-dist}), our experience is already optimally labeled, so relabeling has no~effect.

\begin{figure}[t]
    \centering
    \vspace{-1em}
\begin{minipage}[t]{0.48\textwidth}
\begin{algorithm}[H]
   \caption{Approximate Inverse RL.  \\{\footnotesize When used in HIPI-RL (Alg.~\ref{alg:irl-q-learning}) we only have transitions, so we compute $R_{\psi^{(j)}}^{(i)}$ using Eq.~\ref{eq:opt-relabel-transition} ({\color{blue} blue line}). When used in HIPI-BC (Alg.~\ref{alg:irl-bc}) we have full trajectories, so we compute $R_{\psi^{(j)}}^{(i)}$ using Eq.~\ref{eq:opt-relabel} ({\color{red}red line}).}}
   \label{alg:irl}
   \footnotesize
\begin{algorithmic}
\Function{InverseRL}{$\{(s_t^{(i)}, a_t^{(i)}, s_{t+1}^{(i)}, \psi^{(i)}\}$}
    \For{$j = 1, \cdots, B$} \Comment{task index}
        \For{$i = 1, \cdots, B$} \Comment{state-action index}
        \State {\color{blue} $R_{\psi^{(j)}}^{(i)} \gets \widetilde{Q}(s^{(i)}, a^{(i)}, \psi^{(j)})$ } \Comment{Eq.~\ref{eq:opt-relabel-transition}}
        \State {\color{red} $R_{\psi^{(j)}}^{(i)} \gets \sum_{t'=t} r_{\psi^{(j)}}(s_t^{(i)}, a_t^{(i)})$} \Comment{Eq.~\ref{eq:opt-relabel}}
        \EndFor
        \State $\log Z(\psi^{(j)}) \gets \frac{1}{B} \sum_{i=1}^B e^{R_{\psi^{(j)}}^{(i)}}$
        \EndFor
        \For{$i = 1, \cdots, B$}
        \State $\widetilde{\psi}^{(i)} \sim \Call{Softmax}{R_{\psi^{(1)}}^{(i)} - \log Z(\psi^{(1)}), \cdots}$
    \EndFor
    \State \textbf{return} $\{\widetilde{\psi}^{(i)}\}$
\EndFunction
\end{algorithmic}
\end{algorithm}
\end{minipage}
\hfill
\begin{minipage}[t]{0.50\textwidth}
\begin{algorithm}[H]
   \caption{\textbf{HIPI-RL}: \\Inverse RL for Off-Policy RL}
   \footnotesize
   \label{alg:irl-q-learning}
\begin{algorithmic}
    \While{not converged}
    \State $\{(s_t^{(i)}, a_t^{(i)}, s_{t+1}^{(i)}, \psi^{(i)}\}  \sim \textsc{ReplayBuffer} $
    \State $\{ \widetilde{\psi}^{(i)}\} \gets {\color{blue} \textsc{InverseRL}(\{(s_t^{(i)}, a_t^{(i)}, s_{t+1}^{(i)}, \psi^{(i)})\}) }$
    \State $\widetilde{Q} \gets \Call{MaxEnt RL}{\{(s_t^{(i)}, a_t^{(i)}, s_{t+1}^{(i)}, \widetilde{\psi}^{(i)}\}}$
   \EndWhile
\end{algorithmic}
\end{algorithm}
\vspace{-1.07em}
\begin{algorithm}[H]
   \caption{\textbf{HIPI-BC}: \\Inverse RL for Behavior Cloning}
   \label{alg:irl-bc}
   \footnotesize
\begin{algorithmic}
    \While{not converged}
    \State $\{(s_t^{(i)}, a_t^{(i)}, s_{t+1}^{(i)}, \psi^{(i)}\}  \sim \textsc{ReplayBuffer} $
    \State $\{ \widetilde{\psi}^{(i)}\} \gets {\color{red} \textsc{InverseRL}(\{(s_t^{(i)}, a_t^{(i)}, s_{t+1}^{(i)}, \psi^{(i)})\}) }$
    \State $\theta \gets \theta + \eta \nabla_\theta \sum_i \log \pi_\theta \left(a_t^{(i)} \mid s_t^{(i)}, \widetilde{\psi}^{(i)} \right)$
    \EndWhile
\State \textbf{return} $\pi_\theta$
\end{algorithmic}
\end{algorithm}
\end{minipage}
\vspace{1em}
\end{figure}

\vspace{0.5em}
\section{Using Inverse RL to Accelerate RL}
\vspace{-0.5em}

In this section, we outline a general recipe, Hindsight Inference for Policy Improvement (HIPI), for using inverse RL to accelerate the learning of downstream tasks. Given a dataset of trajectories, we use inverse RL to infer for which tasks those trajectories are optimal. We discuss two options for how to use these relabeled trajectories. One option is to apply off-policy RL on top of these relabeled trajectories. This option generalizes previously-introduced hindsight relabeling techniques~\citep{kaelbling1993learning,andrychowicz2017hindsight}, allowing them to be applied to task distributions beyond goal-reaching. A second option is to apply behavior cloning to the relabeled experience. This option generalizes a number of previous methods, extending variational policy search~\citep{peters2007reinforcement,dayan1997using, levine2013variational, peng2019advantage} to the multi-task setting and extending goal-conditioned imitation learning~\citep{ghosh2019learning,savinov2018semi, lynch2019learning} to arbitrary task distributions.

\vspace{-0.5em}
\subsection{Using Relabeling Data for Off-Policy RL (HIPI-RL)}

Off-policy RL algorithms, such as Q-learning and actor-critic algorithms, represent a broad class of modern RL methods.
These algorithms maintain a replay buffer of previously seen experience, and we can relabel this experience using inverse RL when sampling from the replay buffer. As noted in Section~\ref{sec:special-case}, hindsight experience replay~\citep{andrychowicz2017hindsight}
can be viewed as a special case of this idea. Viewing relabeling as inverse RL, we can extend these methods to general classes of reward functions.

There are many algorithms for inverse methods, and we outline one approximate algorithm that can be efficiently integrated into off-policy RL.
To relabel entire trajectories, we would start by computing the cumulative reward: $R_\psi(s, a) = \sum_{t'=t}r_\psi(s_{t'}, a_{t'})$.
However, most off-policy RL algorithms maintain a replay buffer that stores transitions, rather than entire trajectories. In this case, following Eq.~\ref{eq:opt-relabel-transition}, we instead use the soft Q-function: $R_\psi(s, a) = \tilde{Q}(s_t, a_t, \psi)$.
We approximate
the partition function $\log Z(\psi)$ %
using Monte Carlo samples $(s_i, a_i)$ from within a batch of size $B$:
\vspace{-0.7em}
\begin{equation*}
    \log Z (\psi) = \log \int e^{R_\psi(s, a)} dsda \approx \frac{1}{B} \sum_{i=1}^B e^{R_\psi(s^{(i)}, a^{(i)})}.
\vspace{-1em}
\end{equation*}
We finally sample tasks $\psi^{(i)}$ following Eq.~\ref{eq:opt-relabel}:
\vspace{-0.2em}
\begin{equation*}
    q(\psi^{(i)} \mid s^{(i)}, a^{(i)}) \propto \exp \left(R_\psi(s^{(i)}, a^{(i)}) - \log \tilde{Z}(\psi^{(i)}) \right).
\vspace{-0.2em}
\end{equation*}
We summarize the procedure for relabeling with inverse RL procedure in Alg.~\ref{alg:irl}. The application of relabeling with inverse RL to off-policy RL, which we call HIPI-RL, is summarized in Alg.~\ref{alg:irl-q-learning}.
We emphasize that Alg~\ref{alg:irl} is just one of many methods for performing inverse RL. Alternative methods include gradient-based optimization of the per-sample task, and learning a parametric task-sampler to approximate the optimal relabeling distribution (Eq.~\ref{eq:opt-relabel}). We leave this as future work.

\vspace{-0.3em}
\subsection{Using Relabeled Data for Behavior Cloning}
\vspace{-0.3em}
\label{sec:irl-bc}

We now introduce a second method to use data relabeled with inverse RL to acquire control policies. The idea is quite simple: given arbitrary data, first relabel that data with inverse RL, and then perform task-conditioned behavior cloning. We call this procedure HIPI-BC and summarize it in Alg.~\ref{alg:irl-bc}.
Why should we expect this procedure to work? The intuition is that relabeling with inverse RL makes the joint distribution of tasks and trajectories closer to the target distribution (i.e., it maximizes the multi-task MaxEnt RL objective (Eq.~\ref{eq:kl-joint})). To convert this joint distribution into an actionable representation, we extract the policy implicitly defined by the relabeled trajectories. Behavioral cloning (i.e., supervised learning) does precisely this.

\vspace{-0.2em}
\subsubsection{Relationship to Prior Methods}
\vspace{-0.2em}
Prior work on both goal-conditioned supervised learning, self-imitation learning, and reward-weighted regression can all be understood as special cases. Goal-conditioned supervised learning~\citep{savinov2018semi,ghosh2019learning,lynch2019learning} learns a goal-conditioned policy using a dataset of past experience. For a given state, the action that was actually taken is treated as the correct action (i.e., label) for states reached in the future, and a policy is learned via supervised learning. As discussed in Section~\ref{sec:special-case}, relabeling with the goal actually achieved is a special case of our framework. We refer the reader to those papers for additional evidence for the value of combining inverse RL (albeit a trivial special case) with behavior cloning can effectively learn complex control policies.
Self-imitation learning~\citep{oh2018self} and iterative maximum likelihood training~\citep{liang2016neural} augment RL with supervised learning on a handful of the best previously-seen trajectories, an approach that can be viewed in the inverse RL followed by supervised learning framework. However, because the connection to inverse RL is not made precise, these methods omit the partition function, which may prove problematic when extending these methods to multi-task settings.
Finally, single-task RL methods based on variational policy search~\citep{levine2018reinforcement} and reward-weighted regression~\citep{peters2007reinforcement,peng2019advantage} can also be viewed in this framework. Noting that the optimal relabeling distribution is given as $q(\psi \mid \tau) \propto \exp(R_\psi(\tau) - \log Z(\psi))$, relabeling by sampling from the inverse RL posterior and then performing behavior cloning can be written concisely as the following~objective:
\vspace{-0.5em}
\begin{equation*}
    \int e^{R_\psi(\tau) - \log Z(\psi)} \sum_t \log \pi(a_t \mid s_t, \psi) d \psi d\tau.
\vspace{-0.2em}
\end{equation*}
The key difference between this objective and prior work is the partition function.
The observation that these prior methods are special cases of inverse RL allows us to apply similar ideas to arbitrary classes of reward functions, a capability we showcase in our experiments.

\setlength{\textfloatsep}{20.0pt plus 2.0pt minus 4.0pt}
\setlength{\floatsep}{12.0pt plus 2.0pt minus 2.0pt}

\vspace{-0.6em}
\section{Experiments: Relabeling with Inverse RL Accelerates Learning}
\vspace{-0.6em}

\begin{figure}[t]
    \vspace{-1.5em}
    \centering
    \begin{subfigure}[b]{0.10\linewidth}
        \centering
        \includegraphics[width=\linewidth]{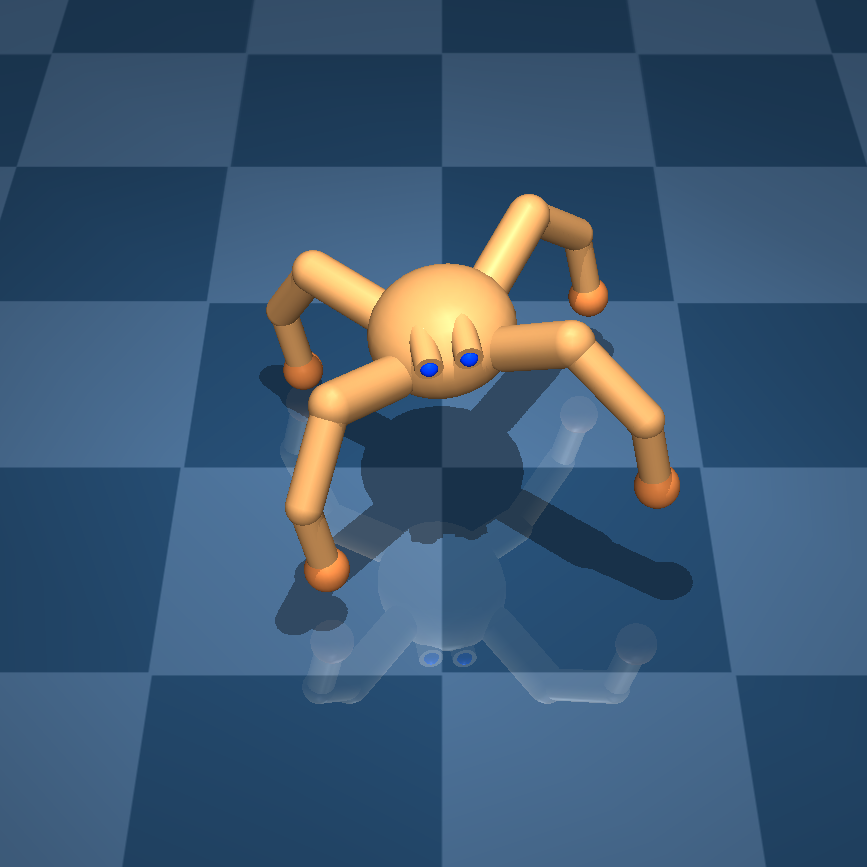}
        \caption{}
    \end{subfigure}%
    ~ 
    \begin{subfigure}[b]{0.10\linewidth}
        \centering
        \includegraphics[width=\linewidth]{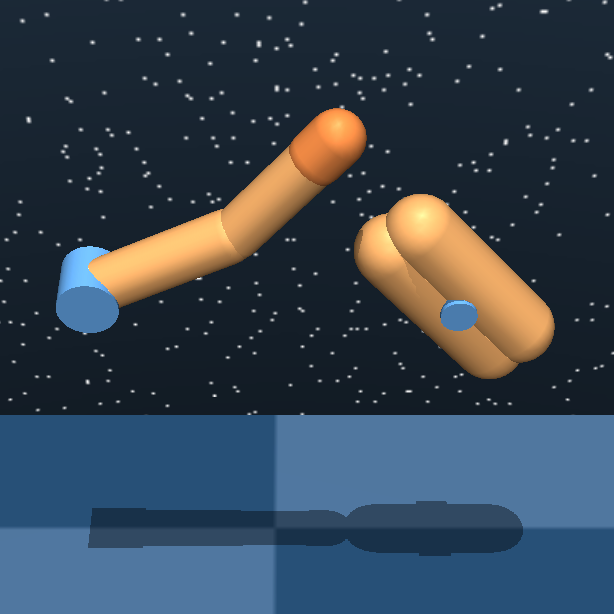}
        \caption{}
    \end{subfigure}%
    ~ 
    \begin{subfigure}[b]{0.10\linewidth}
        \centering
        \includegraphics[width=\linewidth]{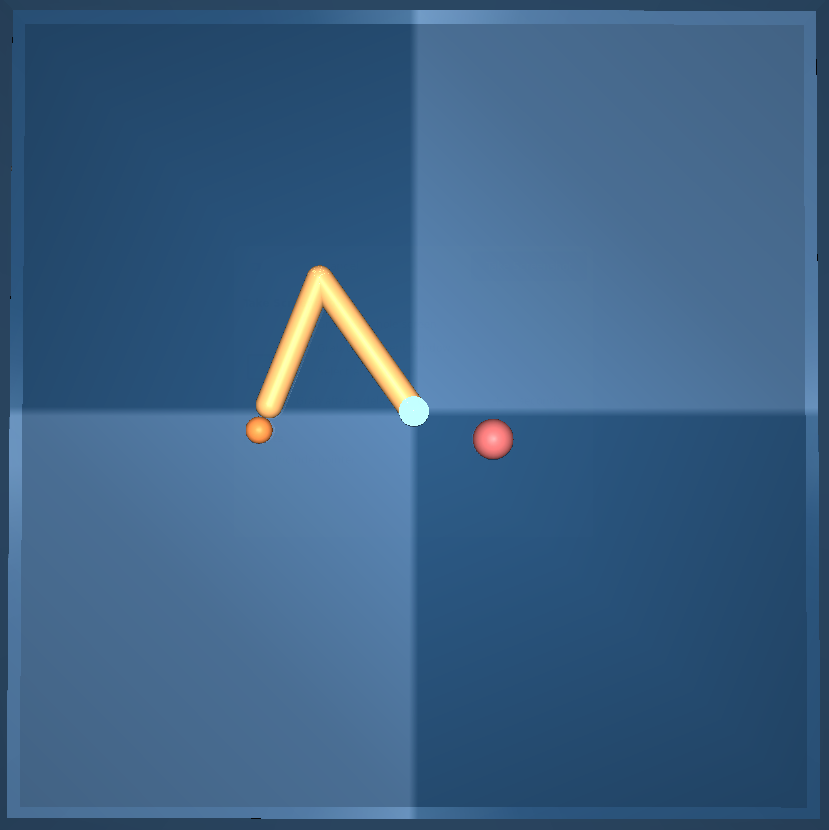}
        \caption{}
    \end{subfigure}%
    ~ 
    \begin{subfigure}[b]{0.1\linewidth}
        \centering
        \includegraphics[width=\linewidth]{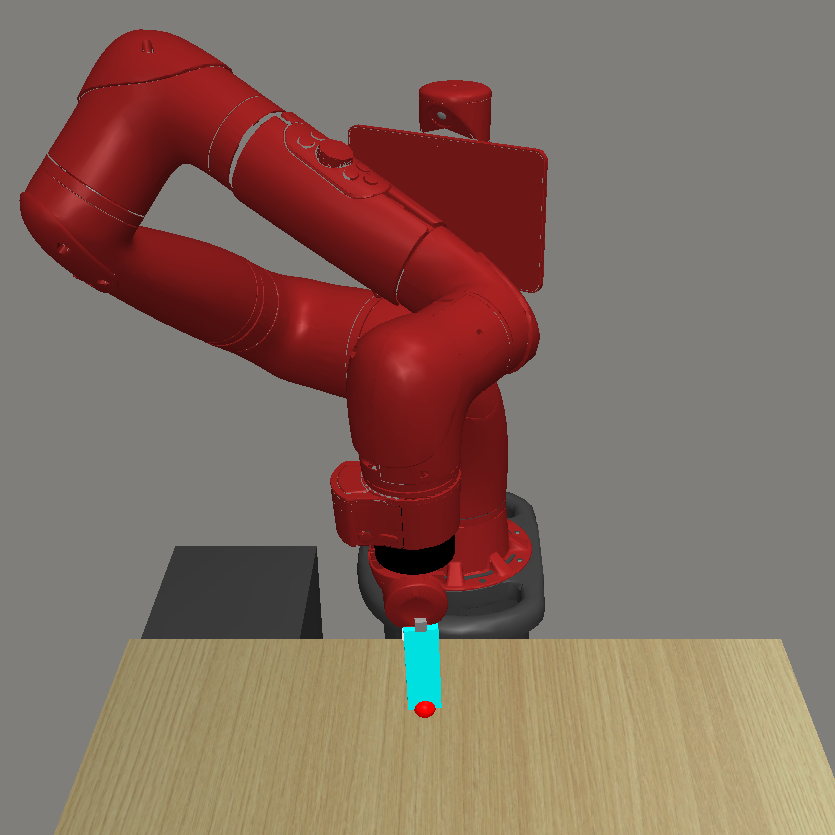}
        \caption{}
    \end{subfigure}%
    ~ 
    \begin{subfigure}[b]{0.1\linewidth}
        \centering
        \includegraphics[width=\linewidth]{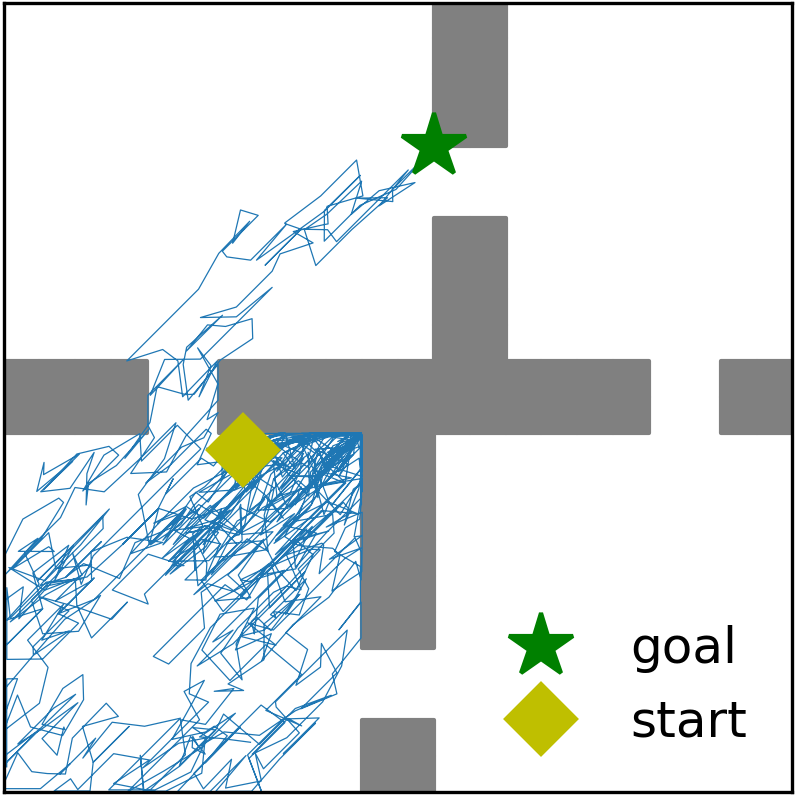}
        \caption{}
    \end{subfigure}%
    ~ 
    \begin{subfigure}[b]{0.1\linewidth}
        \centering
        \includegraphics[width=\linewidth]{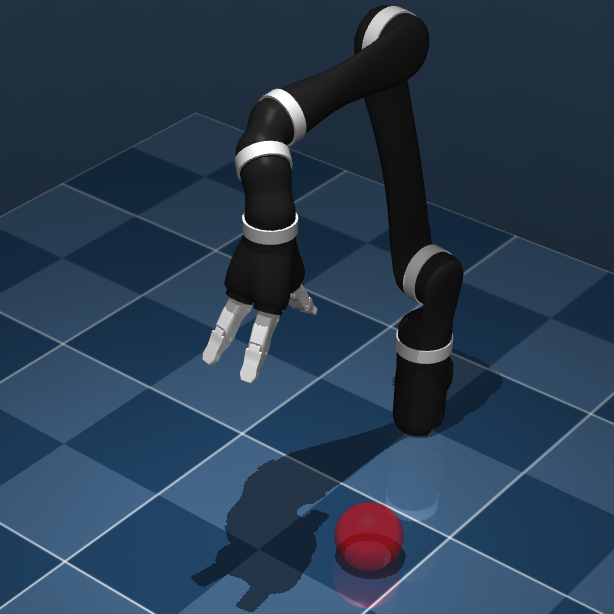}
        \caption{}
    \end{subfigure}%
    ~ 
    \begin{subfigure}[b]{0.1\linewidth}
        \centering
        \includegraphics[width=\linewidth]{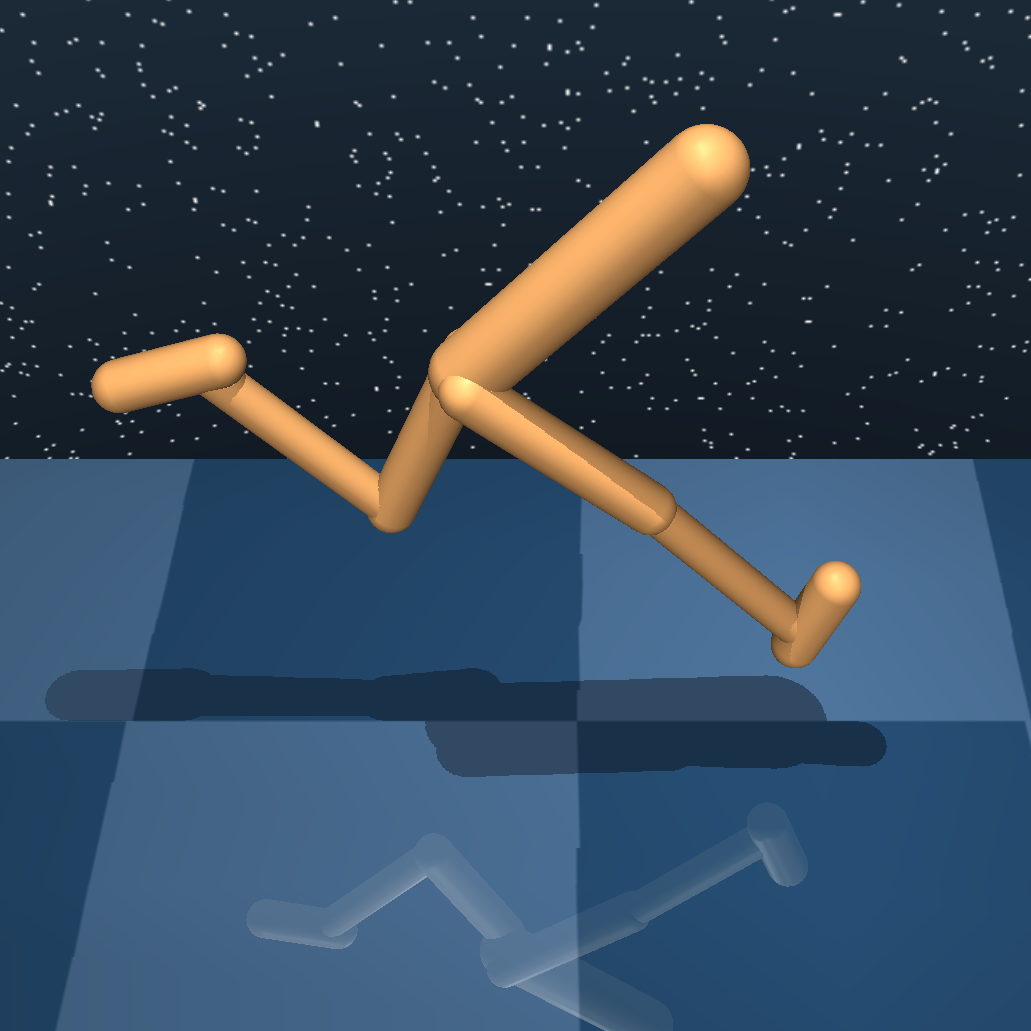}
        \caption{}
    \end{subfigure}%
    ~ 
    \begin{subfigure}[b]{0.1\linewidth}
        \centering
        \includegraphics[width=\linewidth]{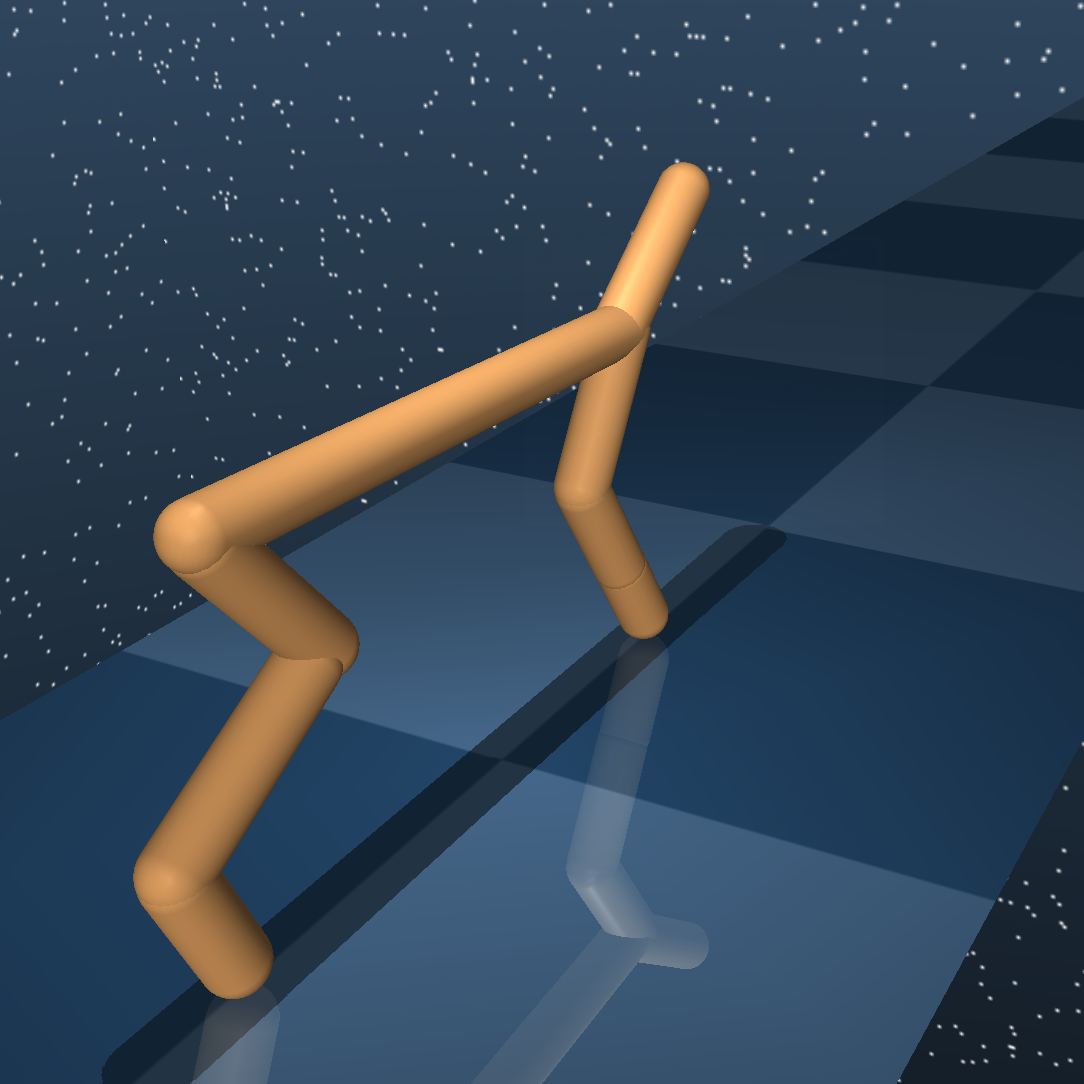}
        \caption{}
    \end{subfigure}%
    ~ 
    \begin{subfigure}[b]{0.1\linewidth}
        \centering
        \includegraphics[width=\linewidth]{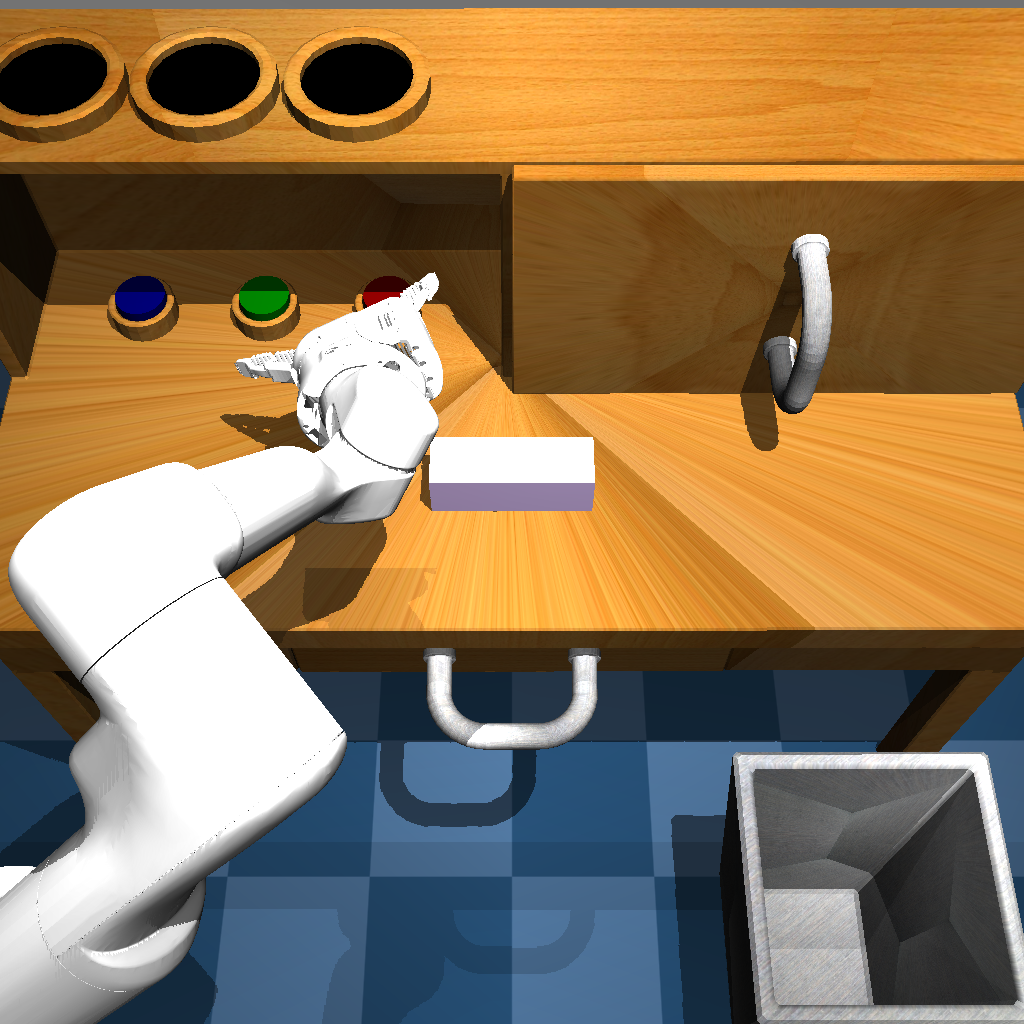}
        \caption{}
    \end{subfigure}
    \vspace{-0.5em}
    \caption{\textbf{Environments for experiments}:
    \emph{(a)} quadruped,
    \emph{(b)} finger,
    \emph{(c)} 2D reacher,
    \emph{(d)} sawyer reach,
    \emph{(e)} 2D navigation
    \emph{(f)} jaco reach,
    \emph{(g)} walker,
    \emph{(h)} cheetah, and
    \emph{(i)} desk manipulation. \label{fig:envs}}
    \vspace{-1.5em}
\end{figure}

Our experiments focus on two methods for using relabeled data: off-policy RL (Alg.~\ref{alg:irl-q-learning}) and behavior cloning (Alg.~\ref{alg:irl-bc}). We evaluate our method on both goal-reaching tasks as well as more general task distributions, including linear combinations of a reward basis and discrete sets of tasks (see Fig.~\ref{fig:envs}).

\vspace{-0.2em}
\subsection{HIPI-RL: Inverse RL for Off-Policy RL}

Our first set of experiments apply Alg.~\ref{alg:irl-q-learning} to domains with varying reward structure, demonstrating how relabeling data with inverse RL can accelerate off-policy RL.

\vspace{-0.2em}
\paragraph{Didactic Example}

\vspace{-0.5em}
\begin{wrapfigure}[18]{r}{0.5\textwidth}
    \vspace{-1.4em}
    \centering
    \includegraphics[width=\linewidth]{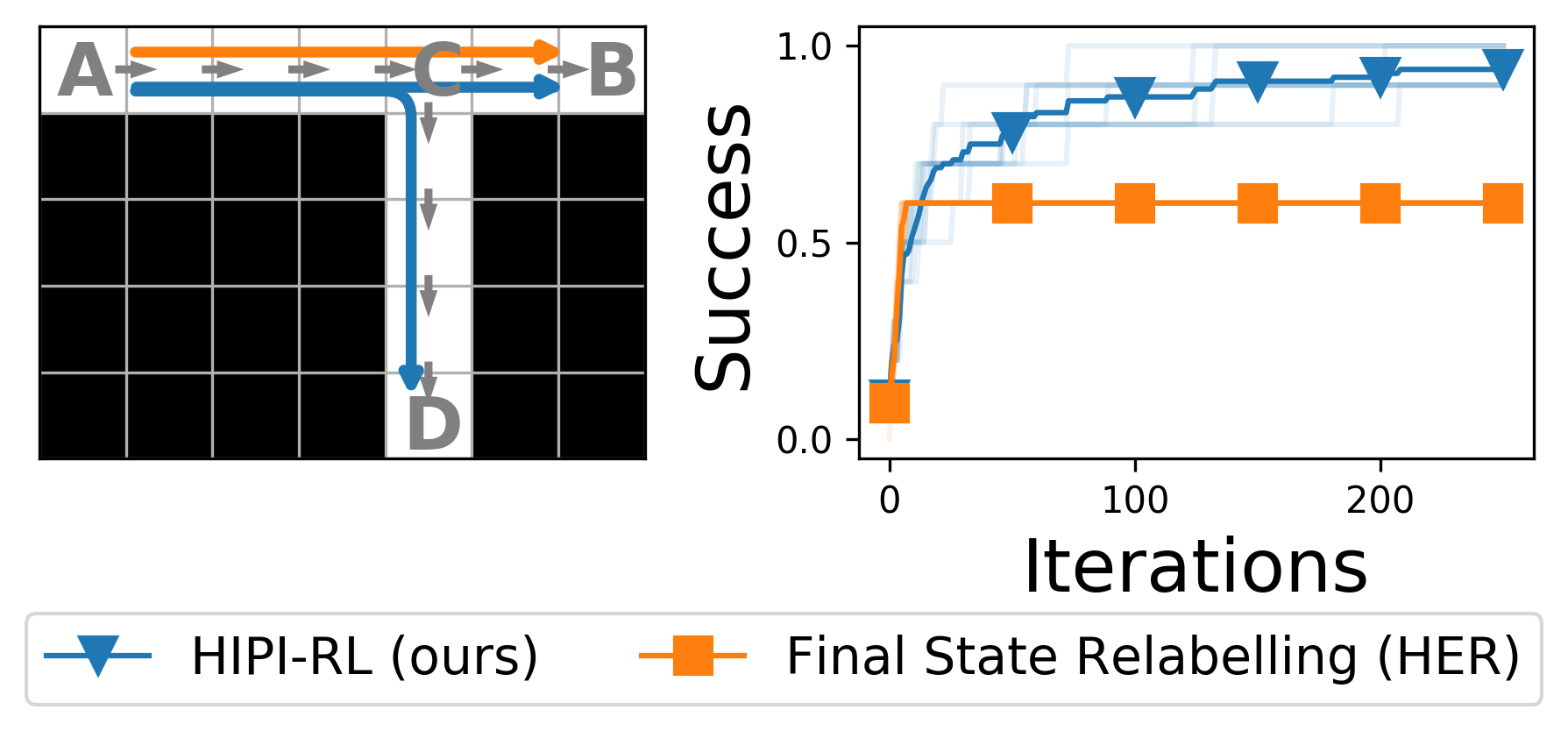}
    \vspace{-1.6em}
     \caption{\footnotesize \textbf{Relabeling stitches crossing trajectories}: \figleft \; A simple gridworld environment, with two observed trajectories $A \rightarrow B$ and $C \rightarrow D$ indicated by grey arrows. Inverse RL identifies both $B$ and $D$ as likely intentions from state $A$ and includes both {\color{custom_blue}$A \rightarrow B$} and {\color{custom_blue}$A \rightarrow D$} in the relabeled data. Final state relabeling (HER) only relabels with the goal actually achieved, corresponding to trajectory {\color{custom_orange}$A \rightarrow B$}. \figright \; We apply Q-learning to both datasets, finding that only relabeling with inverse RL allows the agent to reach all goals. \label{fig:traj-crossing}}
\end{wrapfigure}

We start with a didactic example to motivate why relabeling experience with inverse RL would accelerate off-policy RL.
In the gridworld shown in Fig.~\ref{fig:traj-crossing}, we construct a dataset with two trajectories: $A \rightarrow B$ and $C \rightarrow D$. From state A, inverse RL identifies both $B$ and $D$ as likely intentions, so we include both $A \rightarrow B$ and $A \rightarrow D$ in the relabeled data. Final state relabeling (HER) only uses trajectory $A \rightarrow C$. We then apply Q-learning to both datasets.
to this dataset.
Whereas Q-learning with final state relabeling only succeeds at reaching those goals in the top row, our approach, which corresponds to Q-learning with inverse RL, relabeling succeeds at reaching all goals.
The remainder of this section will show the benefits of relabeling using inverse RL in domains of increasing complexity.

\begin{figure}[t]
    \centering
    \vspace{-2.5em}
    \includegraphics[width=\linewidth]{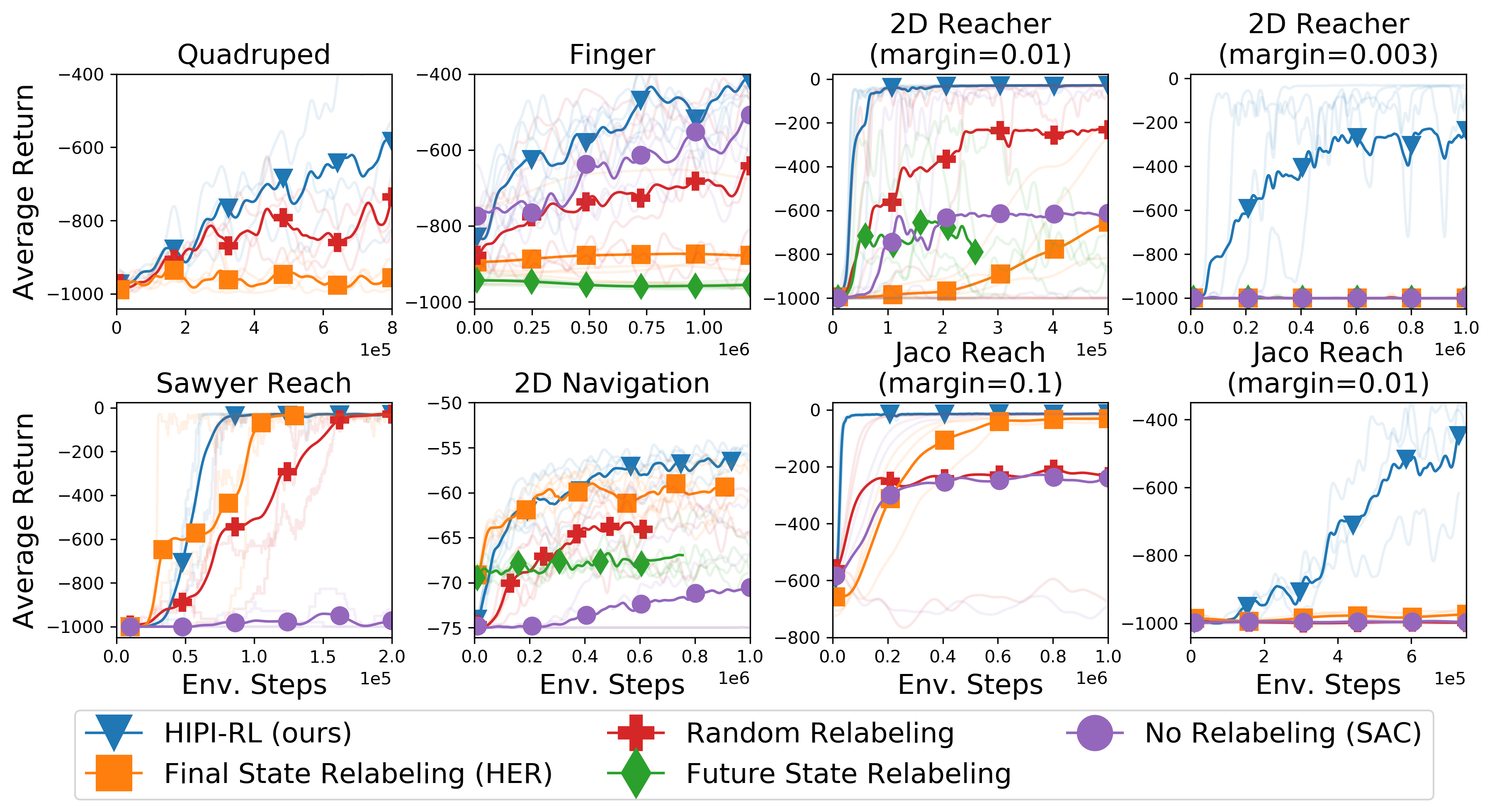}
    \vspace{-1.5em}
    \caption{\footnotesize \textbf{Relabeling for goals-reaching tasks}: On six goal-reaching domains, relabeling with inverse RL (our method) learns faster than with previous relabeling strategies. On extremely sparse versions of two tasks, shown in the right column, only our method learns the tasks.}
    \label{fig:goals}
    \vspace{-1.5em}
\end{figure}

\paragraph{Goal-Reaching Task Distributions} We next apply our method to goal-reaching tasks, where each task $\psi$ corresponds to reaching a different goal state. We used six domains: a quadruped locomotion task, a robotic finger turning a knob, a 2D reacher, a reaching task on the Sawyer robot, a 2D navigation environment with obstacles, and a reaching task on the Jaco robot.
Appendix~\ref{appendix:experiments} provides details of all tasks. We compared our method against four alternative relabeling strategies: relabeling with the final state reached (HER~\citep{andrychowicz2017hindsight}), relabeling with a randomly-sampled task, relabeling with a future state in the same trajectory, and doing no relabeling (SAC~\citep{haarnoja2018soft}).
For tasks where the goal state only specifies certain dimensions of the state, relabeling with the final state and future state requires privileged information indicating to which state dimensions the goal corresponds.

As shown in Fig.~\ref{fig:goals}, relabeling experience with inverse RL (our method) always learns at least as quickly as the other relabeling strategies, and often achieves larger asymptotic reward. While final state relabeling (HER) performs well on some tasks, it is worse than random relabeling on other tasks. We also observe that random relabeling is a competitive baseline, provided that the number of gradient steps is sufficiently tuned.
We conjectured that soft relabeling would be most beneficial in settings with extremely sparse rewards. To test this hypothesis, we modified the reward functions in 2D reacher and Jaco reaching environments to be much sparser. As shown in the far right column on Fig.~\ref{fig:goals}, only soft relabeling is able to make learning progress in this setting.

\begin{figure}[b]
    \centering
    \vspace{-1.5em}
    \includegraphics[width=\linewidth]{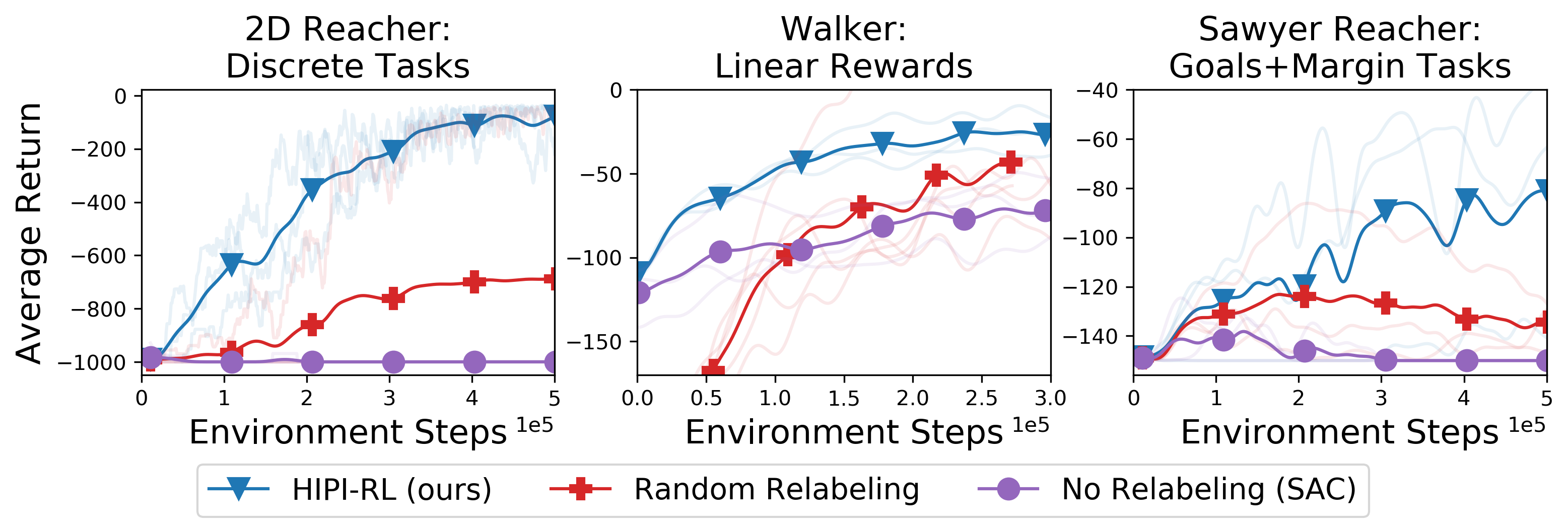}
    \vspace{-1.5em}
    \caption{\footnotesize \textbf{Relabeling for general tasks distributions}: \figleft\; 2D reacher with a discrete set of target end effector positions, \figcenter \; walker with tasks defined as a linear combination of reward terms, and \figright \; sawyer reacher where tasks $\psi = (s_g, m)$ are defined as arriving within $m$ units of goal state $s_g$. On all tasks, relabeling with inverse RL accelerates learning and leads to larger asymptotic reward. Note that existing relabeling strategies are not applicable in this setting.}
    \label{fig:general}
    \vspace{-1.0em}
\end{figure}

\vspace{-0.2em}
\paragraph{More General Task Distributions} Our next experiment demonstrates that, in addition to relabeling goals, inverse RL can also relabel experience for more general tasks distributions.
Our first task distribution is a discrete set of goal states $\psi \in \{1, \cdots, 32\}$ for the 2D reacher environment.
The second task distribution highlights the capability of inverse RL to relabel experience for classes of reward functions defined as linear combinations $r_\psi(s, a) = \sum_{i=1}^d \psi_i \phi_i(s, a)$ of features $\phi(s, a) \in \mathbbm{R}^d$. We use the walker environment, with features corresponding to torso height, velocity, relative position of the feet, and a control cost.
The third task distribution is again a goal reaching task, but one where the task $\phi = (s_g, m)$ indicates both the goal state as well as the desired margin from that goal state. As prior relabeling approaches are not applicable to these general task distributions, we only compared our approach to random relabeling and no relabeling (SAC~\citep{haarnoja2018soft}). As shown in Fig.~\ref{fig:general}, relabeling with inverse RL provides more sample efficient learning in all tasks, and the asymptotic reward is larger than the baselines by a non-trivial amount in two of the three tasks.

\vspace{-0.3em}
\subsection{HIPI-BC: Behavior Cloning on Experience Relabeled with Inverse RL}
\vspace{-0.3em}

\begin{figure}[t]
    \centering
    \vspace{-2em}
    \includegraphics[width=\linewidth]{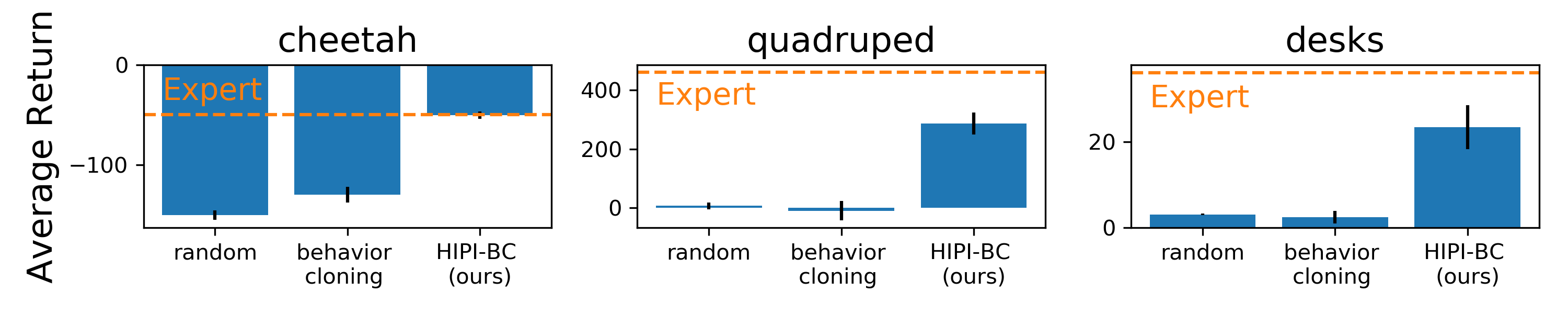}
    \vspace{-2em}
    \caption{\footnotesize \textbf{Behavior cloning on experience relabeled with inverse RL}: We apply our approach to tasks with varying task distributions: \figleft \; goal-reaching tasks on half-cheetah, \figcenter \; linear reward functions on quadruped, and \figright \; discrete tasks on the manipulation environment. Relabeling experience with inverse RL increases reward in all domains. \label{fig:irl-bc}}
    \vspace{-1.5em}
\end{figure}

In this section, we present experiments that use behavior cloning on top of relabeled experience (Alg.~\ref{alg:irl-bc}).
The three domains we use have varying reward structure: (1) half-cheetah with continuous goal velocities; (2) quadruped with linear reward functions; and (3) the manipulation environment with nine discrete tasks. For the half-cheetah and quadruped domains, we collected 1000 demonstrations from a policy trained with off-policy RL. For the manipulation environment, 
\citet{lynch2019learning} provided a dataset of 100 demonstrations for each of these tasks, which we aggregate into a dataset of 900 demonstrations. In all settings, we discarded the task labels, simulating the common real-world setting where experience does not come prepared with task labels.
As shown in Fig.~\ref{fig:irl-bc}, first inferring the tasks with inverse RL and then performing behavioral cloning results in significantly higher final rewards than task-agnostic behavior cloning on the entire dataset, which is no better than random.

\begin{wrapfigure}[12]{r}{0.5\textwidth}
    \centering
    \vspace{-1.0em}
    \includegraphics[width=\linewidth]{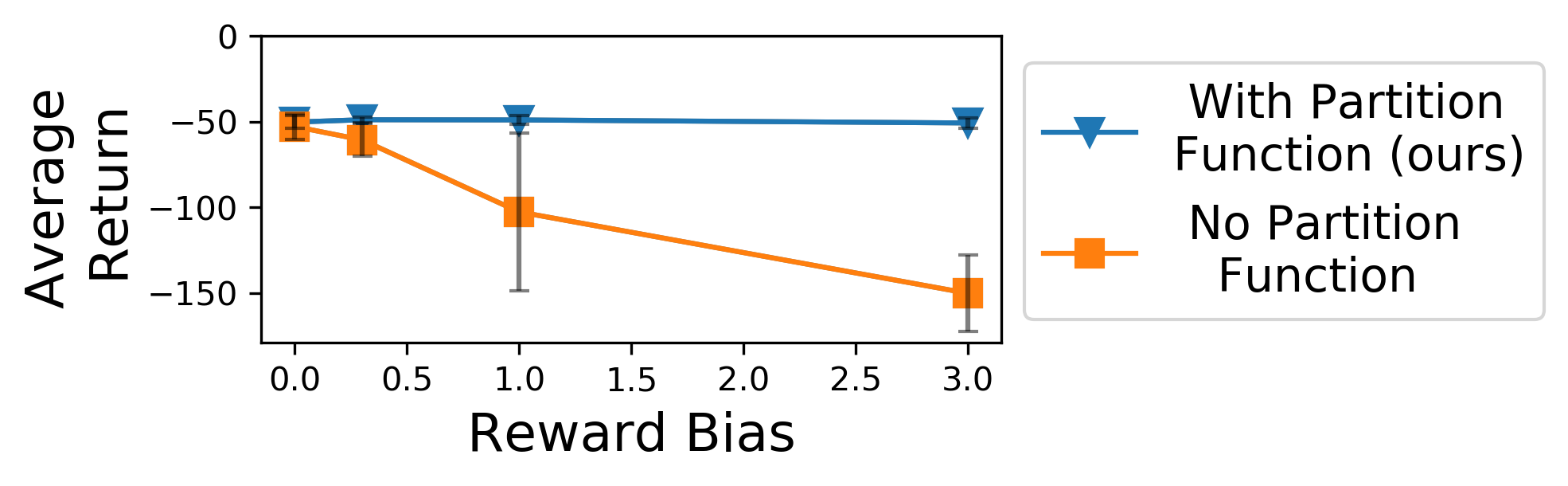}
    \vspace{-1.5em}
    \caption{\footnotesize \textbf{Importance of the partition function}: On the half-cheetah task, we simulated the effect of unnormalized reward functions by adding a constant bias to the first task. Inverse RL normalizes rewards by the partition function. Without this normalization, experience is disproportionately labeled with the first task label, resulting in poor performance during behavior cloning. \label{fig:partition-fn}}
\end{wrapfigure}
Our final experiment demonstrates the importance of the partition function.
On the cheetah domain, we synthetically corrupt the demonstrations by adding a constant bias to the reward for the first task (whichever velocity was sampled first). We then compare the performance of our approach against an ablation that did not normalize by the partition function when relabeling data.
As shown in Fig.~\ref{fig:partition-fn}, using task rewards of different scales significantly degrades the performance of the ablation. Our method, which normalizes the task rewards in the inverse RL step, is not affected by reward scaling.

\vspace{-0.6em}
\section{Discussion}
\vspace{-0.6em}
\label{sec:conclusion}

In this paper, we introduced the idea that hindsight relabeling is inverse RL. We showed that a number of prior works can be understood as special cases of this general framework. The idea that inverse RL might be used to relabel data is powerful because it enables us to extend relabeling techniques to general classes of reward functions. We explored two particular instantiations of this idea, using experience relabeled with inverse RL for off-policy RL and for supervised learning.

We are only scratching the surface of the many ways relabeled experience might be used to accelerate learning. %
For example, the problem of task inference is ever-present in meta-learning, and it is intriguing to imagine explicitly incorporating inverse RL into meta RL.
Broadly, we hope that the observation that inverse RL can be used to accelerate RL will spur research on better inverse RL algorithms, which in turn will provide better RL algorithms.

\vspace{2em}

\paragraph{Acknowledgements}
{\footnotesize We thank Yevgen Chebotar, Aviral Kumar, Vitchyr Pong, and Anirudh Vemula for formative discussions.
We are grateful to Ofir Nachum for pointing out the duality between MaxEnt RL and the partition function, and to Karol Hausman for reviewing an early draft of this paper. We thank Stephanie Chan, Corey Lynch, and Pierre Sermanet for providing the desk manipulation environment.
This research was supported by the Fannie and John Hertz Foundation, NASA, DARPA, US Army, and the National Science Foundation (IIS-1700696, IIS-1700697, IIS1763562, and DGE 1745016). Any opinions, findings and conclusions or recommendations expressed in this material are those of the authors and do not necessarily reflect the views of the National Science Foundation.
}

\clearpage
\appendix

\section{Proof of Lemma~\ref{lemma:improvement}}
\label{appendix:proof}
This section provides a proof of Lemma~\ref{lemma:improvement}.
\begin{proof}
The optimal relabeling distribution can be viewed as an information projection of the joint distribution \mbox{$q_\tau(\psi \mid \tau) q_\tau(\tau)$} onto the target distribution $p(\tau, \psi)$ (Eq.~\ref{eq:target-dist}):
\begin{equation*}
    q_\tau(\psi \mid \tau) q_\tau(\tau) = \min_{q_\tau \in \gQ_\tau} \kl{q_\tau(\psi \mid \tau) q_\tau(\tau, \psi)}{p(\tau, \psi)},
\end{equation*}
where $\gQ = \{q(\tau, \psi) \text{ s.t. } \int q(\tau, \psi) d\psi = q_\tau(\tau)\}$ is the set of all joint distributions with marginal $q_\tau(\tau)$. Note that this set $\gQ$ is closed and convex. We then apply Theorem 11.6.1 from~\citet{cover2006elements}:
\begin{align}
    & \kl{q(\tau, \psi)}{p(\tau, \psi)} \ge \kl{q_\tau(\tau, \psi)}{p(\tau, \psi)} + \kl{q(\tau, \psi)}{q_\tau(\tau, \psi)}. \label{eq:proof}
\end{align}
The second KL divergence on the RHS can be simplified:
\begin{align*}
    \kl{q(\tau, \psi)}{q_\tau(\tau, \psi)} &= \kl{q(\psi \mid \tau) q_\tau(\tau)}{q_\tau(\psi \mid \tau) q_\tau(\tau)} \\
    & = \cancel{\kl{q_\tau(\tau)}{q_\tau(\tau)}} +  \E_{q_\tau} \left[\kl{q(\psi \mid \tau)}{q_\tau(\psi \mid \tau)} \right]
\end{align*}
Substituting this simplification into Eq.~\ref{eq:proof} and rearranging terms, we obtain the desired result.
\end{proof}

\section{Inverse RL on Transitions}
\label{appendix:soft-q}
For simplicity, our derivation of relabeling in Section~\ref{sec:multi-task} assumed that entire trajectories were provided. This section outlines how to do relabeling with inverse RL when we are only provided with $(s, a, s')$ transitions, rather than entire trajectories. This derivation will motivate the use of the \emph{soft} Q-function in Eq.~\ref{eq:opt-relabel-transition}.

In this case, policy distribution $q$ in the MaxEnt RL objective (Eq.~\ref{eq:kl-joint}) is conditioned on the current state and action $(s_t, a_t)$ in addition to the task $\psi$:
\begin{equation}
    \max_{q(\tau, \psi \mid s_t, a_t)} -\kl{q(\tau, \psi \mid s_t, a_t)}{p(\tau, \psi)}.
\end{equation}
Following the derivation in Section~\ref{sec:multi-task}, we expand this objective, using $q(\psi \mid s_t, a_t)$ as our relabeling distribution:
\begin{align}
    \E_{\substack{\psi \sim q(\psi \mid s_t, a_t) \\ \tau \sim  q(\tau \mid \psi, s_t, a_t)}} \bigg[& \sum_{t'=t} r_\psi(s_{t'}, a_{t'}) + \cancel{\log p(s_{t'+1} \mid s_{t'}, a_{t'})} - \log q(a_{t'} \mid s_{t'}, \psi) - \cancel{\log p(s_{t'+1} \mid s_{t'}, a_{t'})} \nonumber \\
    & + p(\psi) - \log q(\psi \mid s_t, a_t) -\log Z(\psi) \bigg]. \label{eq:kl-conditioned}
\end{align}
The expected value of the two summations is the \emph{soft} Q-function for the policy $q(a \mid s, \psi)$:
\begin{align}
    \widetilde{Q}^q(s_t, a_t, \psi) = \E_{\substack{\psi \sim q(\psi \mid s_t, a_t) \\ \tau \sim  q(\tau \mid \psi, s_t, a_t)}} \bigg[ \sum_{t'=t} r_\psi(s_{t'}, a_{t'}) - \log q(a_{t'} \mid s_{t'}, \psi) \bigg]. \label{eq:soft-q}
\end{align}
Substituting Eq.~\ref{eq:soft-q} into Eq.~\ref{eq:kl-conditioned} and ignoring terms that do not depend on $\psi$, we can solve the optimal relabeling distribution:
\begin{equation}
    q(\psi \mid s_t, a_t) \propto p(\psi) e^{\widetilde{Q}^q(s_t, a_t, \psi) - \log Z(\psi)}.
\end{equation}

\section{Experimental Details}
\label{appendix:experiments}

\subsection{Hyperparameters for Off-Policy RL}
Except for the didactic experiment, we used SAC~\citep{haarnoja2018soft} as our RL algorithm, taking the implementation from~\citet{guadarrama2018tf}. This implementation scales the critic loss by a factor of 0.5.
Following prior work~\citep{pong2018temporal}, we only relabeled 50\% of the samples drawn from the replay buffer, using the originally-commanded task the remaining 50\%.
The only hyperparameter that differed across relabeling strategies was the number of gradient updates per environment step. For each experiment, we evaluated each method with values in $\{1, 3, 10, 30\}$ and reported the results of the best hyperparameter in our plots. Perhaps surprisingly, doing random relabeling but simply increasing the number of gradient updates per environment step was a remarkably competitive baseline.
\begin{itemize}
    \setlength\itemsep{0.2em}
    \item Learning Rate: 3e-4 (same for actor, critic, and entropy dual parameter)
    \item Batch Size: 32
    \item Network architecture: The input was the concatenation of the state observation and the task $\psi$. Both the actor and critic networks were 2 hidden layer ReLu networks. The actor output was squashed by a tanh activation to lie within the actor space constraints. There was no activation at the final layer of the critic network, except in the desk environment (see comment below). The hidden layer dimensions were (32, 32) for the 2D navigation environments, (256, 256) for the quadruped and desk environments, and (64, 64) for all other environments.
    \item Discount $\gamma$: 0.99
    \item Initial data collection steps: 1e5
    \item Target network update period: 1
    \item Target network $\tau$: 0.005
    \item Entropy coefficient $\alpha$: We used the entropy-constrained version of SAC~\citep{haarnoja2018apps}, using $-\text{dim}(\gA)$ as the target value, where $\text{dim}(\gA)$ is the action space dimension.
    \item Replay buffer capacity: 1e6
    \item Optimizer: Adam
    \item Gradient Clipping: We found that clipping the gradients to have unit norm was important to get any RL working on the Sawyer and Jaco tasks.
\end{itemize}
To implement final state relabeling, we modified transitions as they were being added to the replay buffer, adding both the original transition and the transition augmented to use the final state as the goal.
To implement future state relabeling, we modified transitions as they were being added to the replay buffer, adding both the original transition and a transition augmented to use one of the next 4 states in the same trajectory as the goal.

\subsection{Hyperparameters for Behavior Cloning Experiments}
To account for randomness in the learning process, we collect at least 200 evaluation episodes per domain; we repeat this experiment for at least 5 random seeds on each domain, and plot the mean and standard deviation over the random seeds.
We used a 2-layer neural network with ReLu activations for all experiments. The hidden layers had size (256, 256). We optimized the network to minimize MSE using the Adam optimizer with a learning rate of 3e-4. We used early stopping, halting training when the validation loss increased for 3 consecutive epochs. Typically training converged in 30 - 50 epochs. We normalized both the states and actions. For the task-conditioned experiments, we concatenated the task vectors to the state vectors.

\subsection{Quadruped Environment}

\begin{wrapfigure}[12]{r}{0.2\textwidth}
\centering
\vspace{-1em}
\includegraphics[width=\linewidth]{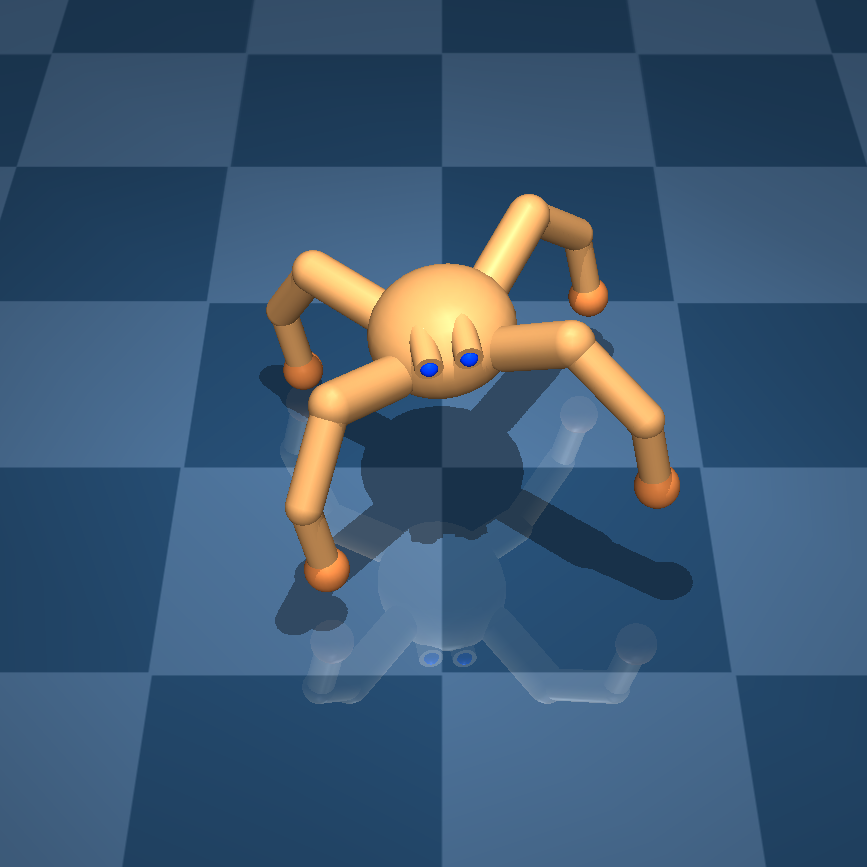}
\caption*{Quadruped}
\end{wrapfigure}
The quadruped was a modified version of the environment from~\citet{abdolmaleki2018maximum}. We modified the initial state distribution so the agent always started upright, and modified the observation space to include the termination signal as part of the observation.
Tasks $\psi \in \mathbbm{R}^2$ were sampled uniformly from the unit circle. Let $s_{\text{XY vel}}$ and $s_{\text{XY pos}}$ indicate the XY velocity and position of the agent.
For the HIPI-RL experiments, we used the following sparse reward function:
\begin{equation*}
    r_\psi(s, a) = \mathbbm{1}(\|s_{\text{XY pos}} - \psi\|_2 \le 0.3) - 1.0,
\end{equation*}
and the episode terminated when $\|s_{\text{XY pos}} - \psi\|_2 \le 0.3$. We also reset the environment after 300 steps if the agent had failed to reach the goal.
For the HIPI-BC experiments, we used the following dense reward function:
\begin{equation*}
    r_\psi(s, a) = s_{\text{XY vel}}^T \psi + 0.1 \|a\|_2^2
\end{equation*}
and episodes were 300 steps long.

\vspace{2em}  %
\subsection{Finger Environment}
\begin{wrapfigure}[9]{r}{0.2\textwidth}
\centering
\vspace{-1.2em}
\includegraphics[width=\linewidth]{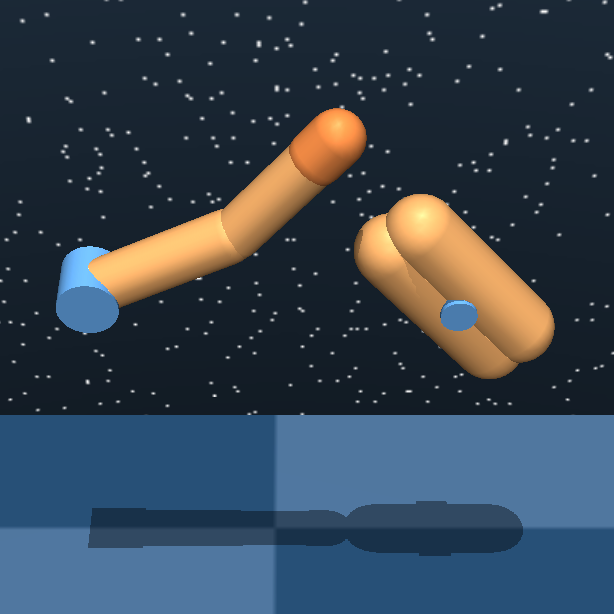}
\caption*{Finger}
\end{wrapfigure}
The finger environment was taken from~\citet{deepmindcontrolsuite2018}. Tasks $\psi$ were sampled using the environment's default goal sampling function. Let $s_{\text{XY}}$ denote the XY position of the knob that the agent can manipulate. The reward function was defined as
\begin{equation*}
    r_\psi(s, a) = \mathbbm{1}(\|s_{\text{XY}} - \psi\|_2 \le 0.01) - 1.0
\end{equation*}
and the episode terminated when $\|s_{\text{XY}} - \psi\|_2 \le 0.01$. We also reset the environment after 1000 steps if the agent had failed to reach the goal.

\subsection{2D Reacher Environment}
\begin{wrapfigure}[9]{r}{0.2\textwidth}
\centering
\vspace{-1.2em}
\includegraphics[width=\linewidth]{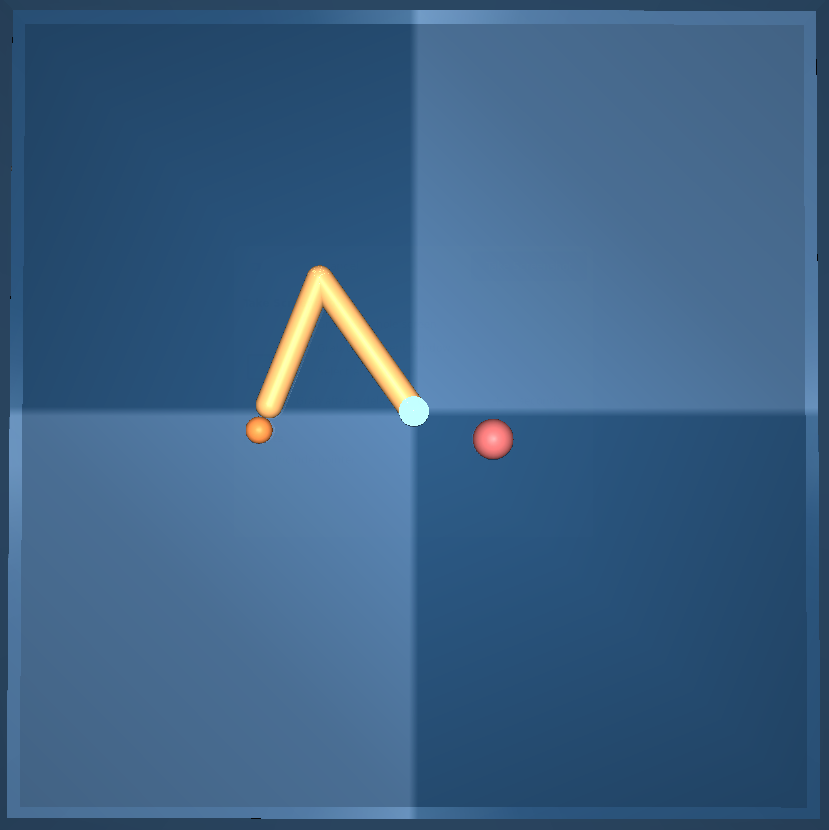}
\caption*{2D Reacher}
\end{wrapfigure}
The 2D reacher environment was taken from~\citet{deepmindcontrolsuite2018}. Let $s_{\text{XY}}$ denote the XY position of the robot end effector. The reward function was defined as
\begin{equation*}
    r_\psi(s, a) = \mathbbm{1}(\|s_{\text{XY}} - \psi\|_2 \le m) - 1.0
\end{equation*}
and the episode terminated when $\|s_{\text{XY}} - \psi\|_2 \le m$, where $m > 0$ is a margin around the goal. We used $m = 0.01$ and $m = 0.003$ in our experiments. We also reset the environment after 1000 steps if the agent had failed to reach the goal.
Tasks were sampled using the environment's default goal sampling function.

\subsection{Sawyer Reach Environment}
\begin{wrapfigure}[11]{r}{0.2\textwidth}
\centering
\vspace{-1.2em}
\includegraphics[width=\linewidth]{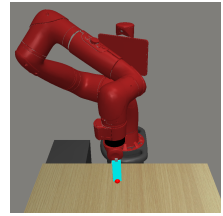}
\caption*{Sawyer Reach}
\end{wrapfigure}
The sawyer reach environment was taken from~\citet{yu2019meta}. Let $s_{\text{XYZ}}$ denote the XYZ position of the robot end effector. The reward function was defined as
\begin{equation*}
    r_\psi(s, a) = \mathbbm{1}(\|s_{\text{XYZ}} - \psi\|_2 \le m) - 1.0
\end{equation*}
and the episode terminated when $\|s_{\text{XY}} - \psi\|_2 \le m$, where $m > 0$ is a margin around the goal. We used $m = 0.01$ and $m = 0.003$ in our experiments. We also reset the environment after 150 steps if the agent had failed to reach the goal.
Tasks were sampled using the environment's default goal sampling function. For the experiment where the task indicator $\psi$ also specified the margin $m$, the margin was sampled uniformly from the interval $[0, 0.1]$.

\subsection{2D Navigation Environment}
\begin{wrapfigure}[9]{r}{0.2\textwidth}
\centering
\vspace{-1.5em}
\includegraphics[width=\linewidth]{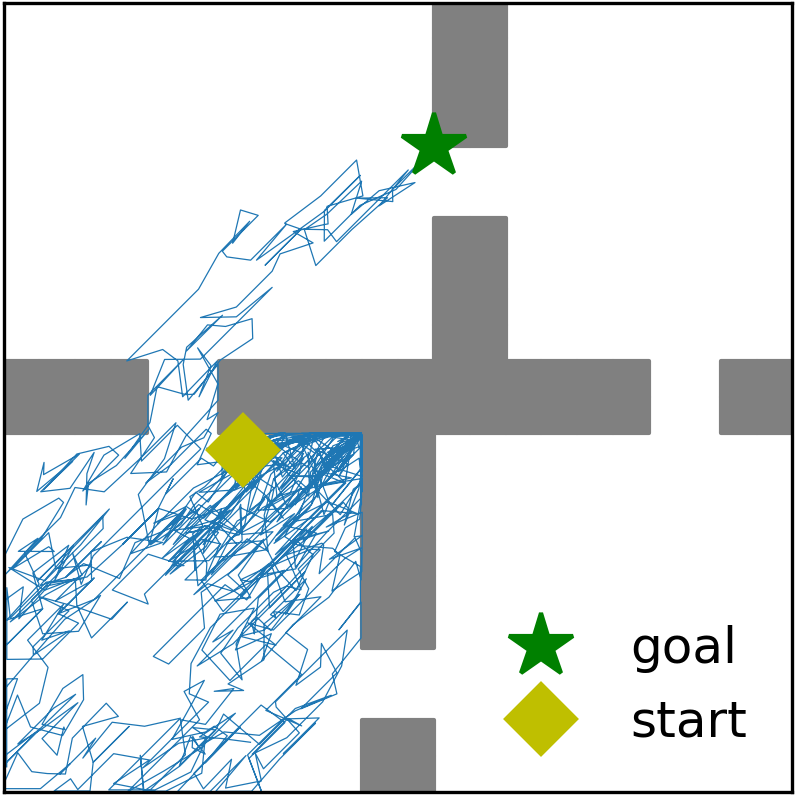}
\caption*{2D Navigation}
\end{wrapfigure}
We used the 2D navigation environment from~\citet{eysenbach2019search}.
The action space is continuous and indicates the desired change of position. The dynamics are stochastic, and the initial state and goal are sampled uniformly at random for each episode. To increase the difficulties of credit assignment and exploration, the agent is always initialized in the lower left corner, and we randomly sampled goal states that are at least 15 steps away.
The layout of the obstacles is taken from the classic FourRooms domain, but dilated by a factor of three.

\subsection{Jaco Reach Environment}
\begin{wrapfigure}[11]{r}{0.2\textwidth}
\centering
\vspace{-1.2em}
\includegraphics[width=\linewidth]{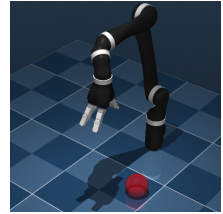}
\caption*{Jaco Reach}
\end{wrapfigure}
We implemented a reaching task using a simulated Jaco robot. Goal states $\psi$ were sampled from uniformly from the interval $[-0.1, 0.1] \times [-0.1, 0.1] \times [0.02, 0.4]$. The agent controlled the velocity of 6 arm joints and 3 finger joints, so the action space was 9 dimensional. The action observation space was 43 dimensional.  
Let $s_{\text{XYZ}}$ denote the XYZ position of the robot end effector. The reward function was defined as
\begin{equation*}
    r_\psi(s, a) = \mathbbm{1}(\|s_{\text{XYZ}} - \psi\|_2 \le m) - 1.0
\end{equation*}
and the episode terminated when $\|s_{\text{XYZ}} - \psi\|_2 \le m$, where $m > 0$ is a margin around the goal. We used $m = 0.1$ and $m = 0.01$ in our experiments. We also reset the environment after 250 steps if the agent had failed to reach the goal.

\subsection{Walker Environment}
\begin{wrapfigure}[11]{r}{0.2\textwidth}
\centering
\vspace{-1.2em}
\includegraphics[width=\linewidth]{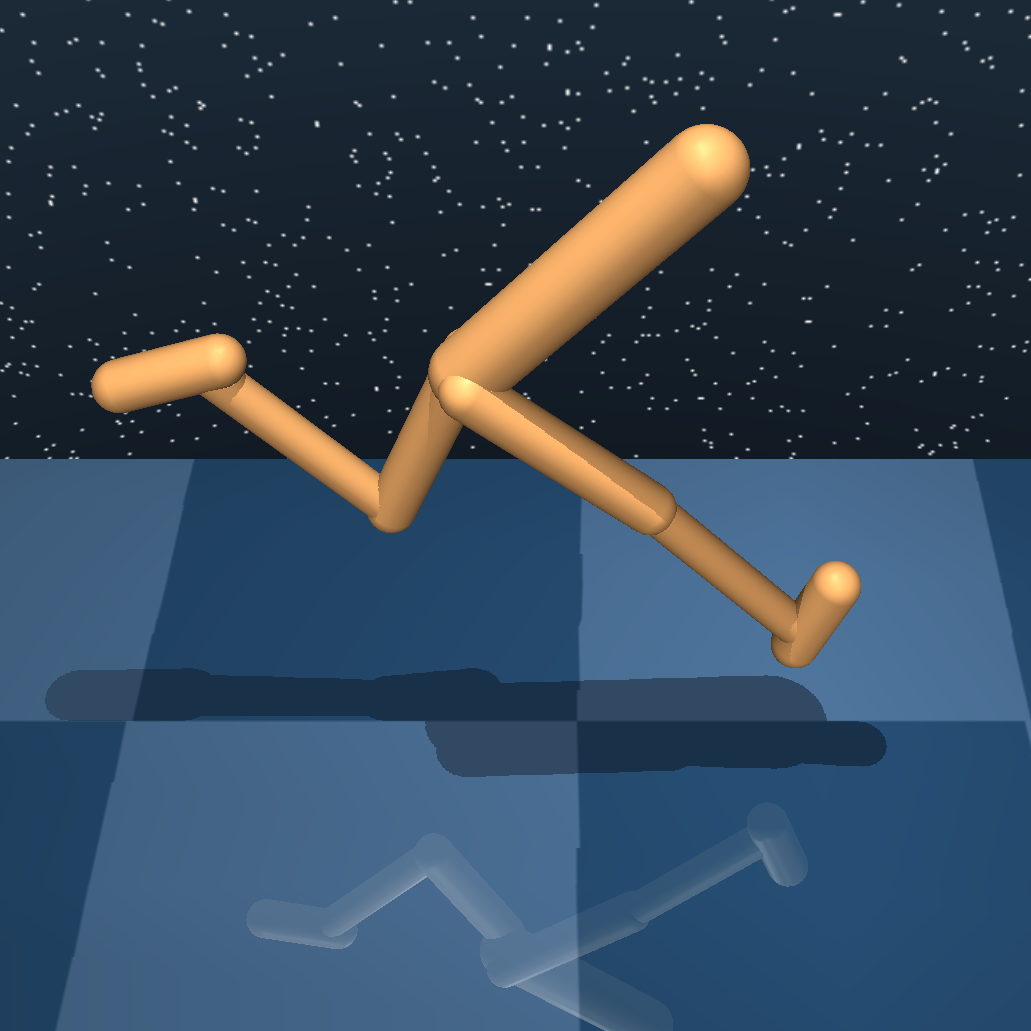}
\caption*{Walker}
\end{wrapfigure}
The walker environment was a modified version of the environment from~\citet{tassa2018deepmind}. We modified the initial state distribution so the agent always started upright, and modified the observation space to include the termination signal as part of the observation. 
For the linear reward function, the features are the torso height (normalized by subtracting 0.5m), velocity along the forward/aft axis, the XZ displacement of the two feet relative to the agent's center of mass (the agent cannot move along the Y axis), and the squared L2 norm of the actions.
The task coefficients $\psi \in \mathbbm{R}^d$ can take on values in the range $[-1, 1]$ for all dimensions, except for the control penalty, which takes on values in $[-1, 0]$.
Episodes were 100 steps long.

\subsection{Half-Cheetah Environment}
\begin{wrapfigure}[9]{r}{0.2\textwidth}
\centering
\vspace{-1.8em}
\includegraphics[width=\linewidth]{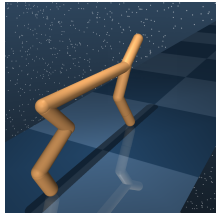}
\caption*{Half-Cheetah}
\end{wrapfigure}
The half-cheetah environment was taken from~\citet{tassa2018deepmind}. 
We define tasks to correspond to goal velocities and use the reward function from~\citet{rakelly2019efficient}:
\begin{equation*}
    r_\psi(s, a) = -|s_{\text{vel}} - \psi| - 0.05 \|a\|_2^2,
\end{equation*}
where $s_{\text{vel}}$ is the horizontal root velocity. Tasks were sampled uniformly $\psi \in [0, 3]$, with units of meters per second.
Episodes were 100 steps long.

\subsection{Desk Environment}
\begin{wrapfigure}[10]{r}{0.2\textwidth}
\centering
\vspace{-1em}
\includegraphics[width=\linewidth]{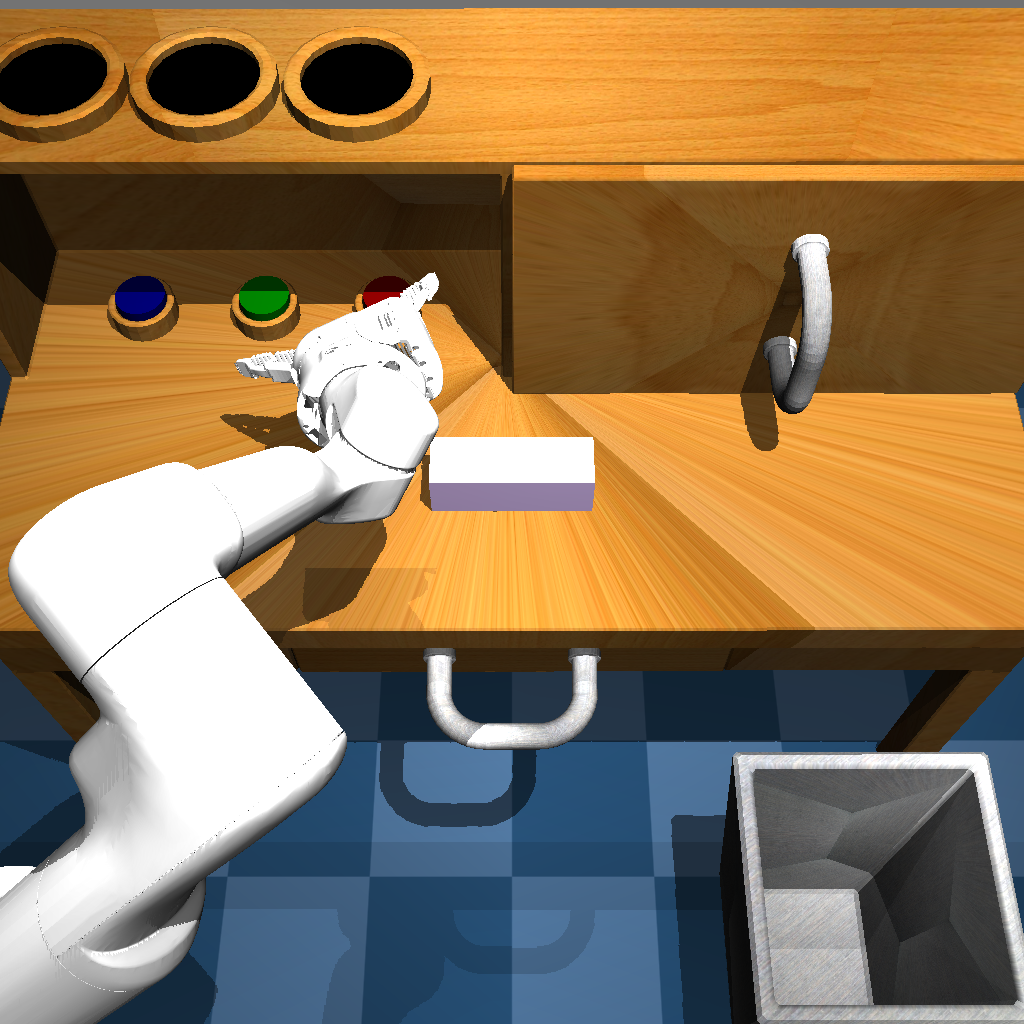}
\caption*{Desk Manipulation}
\end{wrapfigure}
The environment provided by~\citet{lynch2019learning} included 19 tasks. We selected the nine most challenging tasks by looking how often a task was accidentally solved. In the demonstrations for task A, we recorded the average return on the remaining 18 tasks. We chose the nine tasks whose average reward was lowest.  The nine tasks were three button pushing tasks and six block manipulation tasks.

For experiments on this environment, we found that normalizing the action space was crucial. We computed the coordinate-wise mean and standard deviation of the actions from the demonstrations, and modified the environment to implicitly normalize actions by subtracting the mean and dividing by the standard deviation. We clipped the action space to $[-1, +0]$, so the agent was only allowed to command actions within one standard deviation (as measured by the expert demos).

Another trick that was crucial for RL on this environment was clipping the critic outputs. Since the reward was in $[0, 1]$ and the episode length was capped at 128 steps, we squashed the Q-value predictions with a scaled sigmoid to be in the range $[0, 128]$.

\section{Failed Experiments}
\begin{enumerate}
    \item \textbf{100\% Relabeling}: When using inverse RL to relabel data for off-policy RL, we initially relabeled 100\% of samples from the replay buffer, but found that learning was often worse than doing no relabeling at all. We therefore switched to only 50\% relabeling in our experiments. We speculate that retaining some of the originally-commanded goals serves as a sort of hard-negative mining.
    \item \textbf{Coordinate Ascent on Eq.~\ref{eq:kl-joint}}: We attempted to devise an EM-style algorithm that performed coordinate ascent in Eq.~\ref{eq:kl-joint}, alternating between (1) doing MaxEnt RL and (2) relabeling that data and acquiring the corresponding policy via behavior cloning. While we were unable to get this algorithm to outperform standard MaxEnt RL, we conjecture that this procedure would work with the right choice of inverse RL algorithm.
\end{enumerate}

\end{document}